\documentclass{article}

\usepackage[preprint,nonatbib]{neurips_2020}

\usepackage[utf8]{inputenc} % allow utf-8 input
\usepackage[T1]{fontenc}    % use 8-bit T1 fonts
\usepackage{hyperref}       % hyperlinks
\usepackage{url}            % simple URL typesetting
\usepackage{booktabs}       % professional-quality tables
\usepackage{amsmath,amsthm}
\usepackage{amssymb}
\usepackage{bbm}
\usepackage{amsfonts}       % blackboard math symbols
\usepackage{multirow}
\usepackage{nicefrac}       % compact symbols for 1/2, etc.
\usepackage{microtype}      % microtypography

\usepackage[ruled,vlined]{algorithm2e}
\usepackage{subfigure}
\usepackage{graphicx}
\usepackage{caption}
\usepackage{algorithmic}
\usepackage{appendix}
\usepackage{xcolor}

\newcommand{\diag}{\mathop{\mathrm{diag}}}
\newtheorem{lemma}{Lemma}

\title{Low-Rank Autoregressive Tensor Completion for Multivariate Time Series Forecasting}

\author{%
  Xinyu Chen \\
  McGill University \\
  \texttt{chenxy346@gmail.com}\\
  \And
  Lijun Sun \\
  McGill University\\
  \texttt{lijun.sun@mcgill.ca}\\
}

\begin{document}

\maketitle

\begin{abstract}
Time series prediction has been a long-standing research topic and an essential application in many domains. Modern time series collected from sensor networks (e.g., energy consumption and traffic flow) are often large-scale and incomplete with considerable corruption and missing values, making it difficult to perform accurate predictions. In this paper, we propose a low-rank autoregressive tensor completion (LATC) framework to model multivariate time series data. The key of LATC is to transform the original multivariate time series matrix (e.g., sensor$\times$time point) to a third-order tensor structure (e.g., sensor$\times$time of day$\times$day) by introducing an additional temporal dimension, which allows us to model the inherent rhythms and seasonality of time series as global patterns. With the tensor structure, we can transform the time series prediction and missing data imputation problems into a universal low-rank tensor completion problem. Besides minimizing tensor rank, we also integrate a novel autoregressive norm on the original matrix representation into the objective function. The two components serve different roles. The low-rank structure allows us to effectively capture the global consistency and trends across all the three dimensions (i.e., similarity among sensors, similarity of different days, and current time \textit{v.s.} the same time of historical days). The autoregressive norm can better model the local temporal trends. Our numerical experiments on three real-world data sets demonstrate the superiority of the integration of global and local trends in LATC in both missing data imputation and rolling prediction tasks.
\end{abstract}

\section{Introduction}

Time series prediction serves as the foundation for a wide range of real-world applications and decision-making processes. Although the field of time series analysis has been developed for a long time, traditional time series models (e.g., autoregressive (AR), ARIMA, exponential smoothing) mainly focus on parametric models for small-scale problems \cite{hyndman2018forecasting}. However, the structure and properties of emerging ``big'' time series data have posed new challenges \cite{faloutsos2018forecasting}. In particular, modern time series data collected from field applications and sensor networks are often large-scale (e.g., time series from thousands of sensors), high-dimensional (e.g., matrix/tensor-variate sequences \cite{xiong2010temporal,jing2018high}), and incomplete (even sparse) with considerable corruption and missing values. A critical challenge is to perform efficient and reliable predictions for large-scale time series with missing values \cite{yu2016temporal}.

The fundamental of modeling of modern large-scale, high-dimensional time series is to effectively characterize the complex dependencies and correlations across different dimensions. Spatiotemporal data is an excellent example with complex dependencies/correlations across both spatial and temporal dimensions \cite{bahadori2014fast}. Taking traffic flow time series collected from a network of sensors as an example, we often observe clear spatial consistency (e.g., nearby sensors generate similar readings) and both long-term and short-term temporal trends and correlations \cite{li2015trend}. To model both long-term and short-term correlations, various neural sequence models have been developed \cite{yu2017long,lai2018modeling,li2019enhancing,sen2019think}. However, these models cannot effectively deal with the missing data problem, and most of them requiring performing imputation as a preprocessing step, which may introduce potential bias. How to incorporate long-term/short-term patterns and sensor correlation in the presence of missing data remains an important research question.

To address data incompleteness, dimensionality reduction methods such as matrix/tensor factorization have been applied to model multivariate and high-order (matrix/tensor-variate) time series data \cite{xiong2010temporal,yu2016temporal,de2017tensorcast,jing2018high,sun2019bayesian}. To acquire prediction power, it is essential for these factorization-based models to impose certain time series structures on the latent layer. However, real-world time series data often have complex temporal structures beyond the simple AR model, exhibiting patterns at different scales and resolutions (weekly, daily, and hourly). For example, human behavior related time series (e.g., household energy consumption and highway traffic flow data) often simultaneously demonstrate long-term patterns at the daily and weekly levels and short-term perturbations. Clearly, these short-term and long-term patterns are critical to prediction tasks. Potential solutions to accommodate both of them are to introduce kernel structures in Gaussian process \cite{roberts2013gaussian} and to use the Hankel transformation following singular spectrum analysis and Hankel structured low-rank models \cite{golyandina2001analysis,markovskylow}; however, the computational cost makes it challenging for large (number of sensors) and long (number of time points) multivariate time series data.

\begin{figure}[!ht]
  \centering
  \includegraphics[scale=0.75]{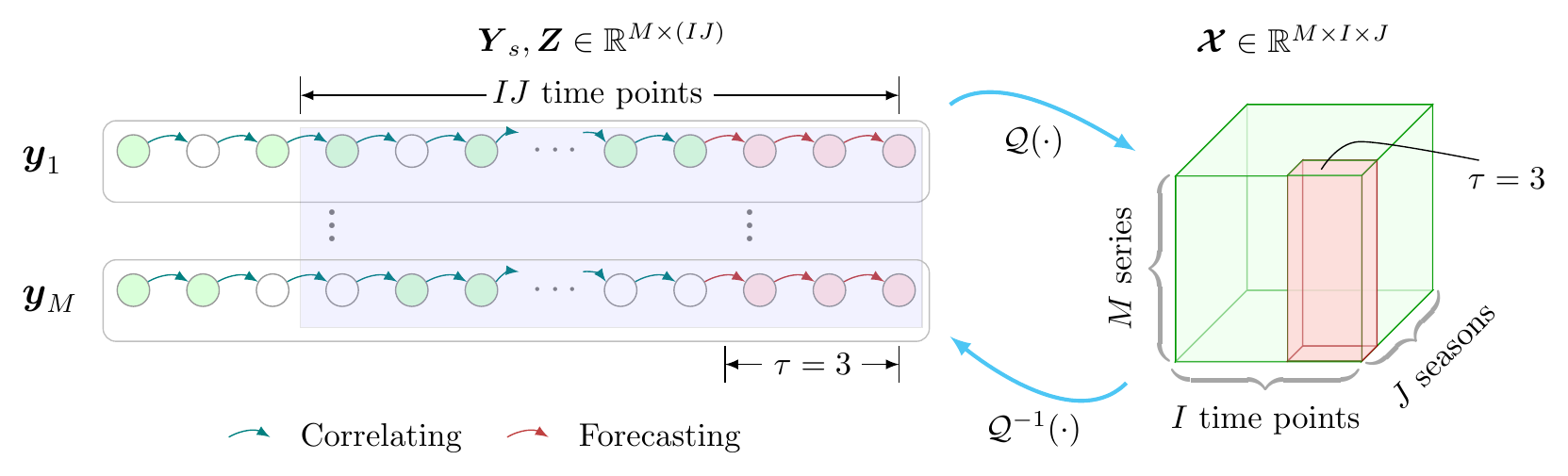}
  \caption{Illustration of the proposed LATC imputer/predictor with a prediction window $\tau$ (green nodes: observed values; white nodes: missing values; red nodes/panel: prediction; blue panel: training data to construct the tensor). For example, if $\boldsymbol{Y}$ captures hourly traffic flow and the goal is perform prediction for the next $\tau=3$ hours, we can set $I=24$ and choose $J$ accordingly.}
  \label{fig:key}
\end{figure}

In this paper, we propose a low-rank autoregressive tensor completion (LATC) framework to model large-scale multivariate time series with missing values. In LATC, the original multivariate time series matrix is transformed to a third-order tensor based on the most important seasonality, and thus both missing data imputation and future value prediction problems can be naturally translated to a universal tensor completion problem (see Figure~\ref{fig:key}). To achieve better prediction power, we use tensor nuclear norm minimization and truncated nuclear norm minimization to preserve the long-term/global trends. We then define a new autoregressive norm on the original matrix representation to characterize short-term/local trends. With this approach, all the observed data in tensor $\boldsymbol{\mathcal{X}}$ will contribute to the final prediction (i.e., the red panel in Figure~\ref{fig:key}). We evaluate LATC on three real-world data sets for both missing data imputation and rolling prediction tasks, and compare it with several state-of-the-art approaches. Our numerical experiments show encouraging performance of LATC, suggesting that the model can effectively capture both global and local trends in time series data.

%\cite{shi2020block}
%\cite{wang2020hkmf}
%\cite{butcher2017simple}
%\cite{li2019tensor}
%\cite{li2017low}

\section{Related work}

In this paper, we focus on developing low-rank models for large-scale multivariate time series data in the presence of missing values. Essentially, there are three types of approaches for this problem.

\textbf{Temporal factorization } Low-rank completion is a popular technique for collaborative filtering and missing data imputation. Some recent studies have used low-rank models for multivariate and high-order time series data \cite{xiong2010temporal,yu2016temporal,jing2018high,sen2019think,sun2019bayesian}. Essentially, these models require a well-designed generative mechanism on the temporal latent layer to achieve smoothing and to harness prediction power. For example, AR models are used to regularize factor matrix and core tensors in \cite{yu2016temporal,sun2019bayesian} and \cite{jing2018high}. However, these models essentially only capture the global consistency/similarity among different time series, but they cannot effectively accommodate the global trends (daily/weekly patterns) on the temporal dimension. Therefore, a critical challenge is to design an effective temporal model, since AR might be too limited to capture periodic pattern at different scales. %To better capture periodic patterns, traditional time series models often incorporate seasonality components (e.g., Seasonal ARIMA model \cite{hyndman2018forecasting}).

\textbf{Hankel/delay embedding } Singular spectrum analysis (SSA) and Hankel structured low-rank completion are powerful approaches for time series analysis \cite{golyandina2001analysis,markovskylow}. They are model-free approaches to detect spectral patterns at different scales in time series data. Essentially, SSA applies singular value decomposition (SVD) on the Hankel matrix obtained from the original univariate time series, and then uses those principle components to analyze the time series. The default SSA model is for univariate time series but it can be easily extended to the multivariate case. This approach can accomplish both missing data imputation and prediction tasks by performing low-rank completion on the Hankel matrix \cite{chen2014robust,gillard2018structured,agarwal2018model,yokota2018missing,zhang2019correction,shi2020block}. However, such model is computationally expensive to model the Hankel matrices/tensors. It might be infeasible to deal with large and long multivariate time series.

\textbf{Tensor representation } Another approach is to fold a time series matrix into a tensor by introducing an additional ``season'' dimension (e.g., a third-order tensor with a ``day'' dimension). In fact, many real-world time series data resulted from human behavior/activities (e.g., traffic flow, customer demand, electricity consumption) exhibit both long-term and short-term patterns. Recent studies have used the tensor representation [\textit{sensor}$\times$\textit{day}$\times$\textit{time of day}] to capture the patterns (e.g., \cite{figueiredo2013electrical,tan2016short,li2019tensor,chen2020nonconvex}). The tensor representation also offers prediction ability by performing tensor completion \cite{tan2016short,li2019tensor}. The tensor representation not only preserves the dependencies among sensors but also provides a new alternative to capture both local and global temporal patterns. These models have shown superiority over matrix-based models in missing data imputation tasks \cite{chen2020nonconvex}.

\section{Preliminaries}

\paragraph{Notations} We use boldface uppercase letters to denote matrices, e.g., $\boldsymbol{X}\in\mathbb{R}^{M\times N}$, boldface lowercase letters to denote vectors, e.g., $\boldsymbol{x}\in\mathbb{R}^{M}$, and lowercase letters to denote scalars, e.g., $x$. Given a multivariate time series matrix $\boldsymbol{X}\in\mathbb{R}^{M\times N}$ ($M$ sensors over $N$ time points), we use $\boldsymbol{X}_{[:t]}\in\mathbb{R}^{M\times t}$ and $\boldsymbol{X}_{[t+1:]}\in\mathbb{R}^{M\times (N-t)}$ to denote the submatrices of $\boldsymbol{X}$ that consist of the first $t$ columns and the last $N-t$ columns, respectively. Let $\boldsymbol{X}_{[t_1:t_2]}\in\mathbb{R}^{M\times (t_2-t_1+1)}$ denote the submatrix of $\boldsymbol{X}$ formed by columns $t_1$ to $t_2$ and $\boldsymbol{x}_t$ denote the vector at time $t$. We denote the $(m,n)$th entry in $\boldsymbol{X}$ by $x_{m,n}$. The Frobenius norm of $\boldsymbol{X}$ is defined as $\|\boldsymbol{X}\|_{F}=\sqrt{\sum_{m,n}x_{m,n}^{2}}$. The nuclear norm (NN) is defined as  $\|\boldsymbol{X}\|_{*}=\sum_{i=1}^{\min\{M,N\}}\sigma_i$, where $\sigma_i$ denotes the $i$th largest singular value of $\boldsymbol{X}$. The truncated nuclear norm (TNN) of $\boldsymbol{X}$ is defined as the sum of the $\left(\min\{M,N\}-\theta\right)$ smallest singular values, i.e., $\|\boldsymbol{X}\|_{\theta,*}=\sum_{i=\theta+1}^{\min\{M,N\}} \sigma_i$. The $\ell_2$-norm of $\boldsymbol{x}$ is defined as $\|\boldsymbol{x}\|_{2}=\sqrt{\sum_{m}x_{m}^{2}}$. We denote a third-order tensor by $\boldsymbol{\mathcal{X}}\in\mathbb{R}^{M\times I\times J}$ and the $k$th-mode ($k=1,2,3$) unfolding of $\boldsymbol{\mathcal{X}}$ by $\boldsymbol{\mathcal{X}}_{(k)}$ \cite{kolda2009tensor}. Correspondingly, we define a folding operator that converts a matrix to a third-order tensor in the $k$th-mode as $\operatorname{fold}_{k}(\cdot)$; thus, we have $\operatorname{fold}_{k}(\boldsymbol{\mathcal{X}}_{(k)})=\boldsymbol{\mathcal{X}}$.

\paragraph{Low-rank matrix/tensor completion} The low-rank matrix completion (LRMC) model imposes an underlying low-rank structure to recover an incomplete matrix. Given a partially observed matrix $\boldsymbol{Y}\in\mathbb{R}^{M\times N}$ with an index set $\Omega$ of observed entries, LRMC can be formulated as:
\begin{equation}
    \begin{aligned}
    \min _{\boldsymbol{X}}~&\operatorname{rank}(\boldsymbol{X}) \\ \text { s.t.}~& \mathcal{P}_{\Omega}(\boldsymbol{X})=\mathcal{P}_{\Omega}(\boldsymbol{Y}), \\
    \end{aligned}
    \label{equ:lrmc}
\end{equation}
where $\boldsymbol{X}$ is the recovered matrix. The symbol $\operatorname{rank}(\cdot)$ denotes the rank of a given matrix. The operator $\mathcal{P}_{\Omega}:\mathbb{R}^{M\times N}\mapsto\mathbb{R}^{M\times N}$ is an orthogonal projection supported on $\Omega$:
\begin{equation} \notag
    [\mathcal{P}_{\Omega}(\boldsymbol{X})]_{m,n}=\left\{\begin{array}{ll}
    x_{m,n},     & \text{if $(m,n)\in\Omega$,}  \\
    0,     & \text{otherwise}. \\
    \end{array}\right.
\end{equation}
Problem \eqref{equ:lrmc} is NP-hard. A convex relaxation is to use $\|\boldsymbol{X}\|_{*}$ as a surrogate for the rank function \cite{recht2010guaranteed}. Low-rank tensor completion (LRTC) extends LRMC to higher-order tensors. Ref.~\cite{liu2013tensor} defines the tensor nuclear norm as the weighted sum of NNs of all the unfolded matrices, i.e., $\|\boldsymbol{\cal{X}}\|_* = \sum_k \alpha_k \|\boldsymbol{\cal{X}}_{(k)}\|_*$, where $\alpha_k$s are non-negative weight parameters with $\sum_{k}\alpha_k=1$.

\paragraph{Autoregressive model of time series} Autoregressive model is a standard statistical model for time series. Given a time series matrix $\boldsymbol{X}\in\mathbb{R}^{M\times N}$, the vector autoregressive (VAR) model gives
\begin{equation}
    \boldsymbol{x}_t = \sum_{i} \boldsymbol{A}_i \boldsymbol{x}_{t-h_i} + {\epsilon}_{t},
    \label{auto_regressive}
\end{equation}
where $\mathcal{H}=\{h_1,\ldots,h_d\}$ is a set of time lags, $\boldsymbol{A}_{i}$ is a coefficients matrix for lag $h_i$, and $\epsilon_{mt}$s are Gaussian noises. VAR can capture the dependencies among different time series, but in the meanwhile it has a large number of parameters $M\times M\times d$. We can also model each individual time series follows an independent autoregressive model as ${x}_{m,t}=\sum_{i}{a}_{m,i}{x}_{m,t-h_i}+{\epsilon}_{m,t}$, and this reduces the coefficients to a matrix $\boldsymbol{A}=[a_{m,i}]\in\mathbb{R}^{M\times d}$.

\section{Low-rank autoregressive tensor completion}

In this section, we introduce the low-rank autoregressive tensor completion (LATC) framework to impute missing values and predict future values of a multivariate time series matrix $\boldsymbol{Y}$. The setting of LATC is essentially the same as in TRMF \cite{yu2016temporal} and BTMF \cite{sun2019bayesian}. The two models introduce an autoregressive regularizer (on the temporal factor matrix $\boldsymbol{X}$) to characterize temporal dynamics when factorizing matrix $\boldsymbol{Y}\approx \boldsymbol{F}\boldsymbol{X}$. The learned autoregressive regularizer enables us perform prediction on the temporal factor matrix $\boldsymbol{X}$ to get $\boldsymbol{X}_{\text{new}}$, and then the final prediction can be obtained by $\boldsymbol{Y}_{\text{new}} = \boldsymbol{F}\boldsymbol{X}_{\text{new}}$. A fundamental challenge in  factorization-based time series models is to design the structural model to effectively capture both long-term and short-term temporal dependencies.

The key idea of the proposed LATC model is to transform the \textbf{matrix-based prediction/imputation} problem to a universal low-rank \textbf{tensor completion} problem (see the illustration in Figure~\ref{fig:key}). Our main motivation to transform the time series matrix to a tensor is that many real-world time series, such as traffic flow and energy consumption data, are characterized by both long-term/global trends and short-term/local trends \cite{li2015trend}. The long-term trends refer to certain periodic, seasonal, and cyclical patterns. For example, traffic flow data over 24 hours on a typical weekday often shows a systematic ``M'' shape resulted from travelers' behavioral rhythms, with two peaks during morning and evening rush hours \cite{lai2018modeling}. The pattern also exists at the weekly level with substantial differences from weekdays to weekends. The short-term trends capture certain temporary volatility/perturbation that deviates from the global patterns (e.g., due to incident/event). The short-term trends seem more ``random'', but they are common and ubiquitous in reality. LATC leverages both global and local patterns
by using a tensor structure. As shown in Figure~\ref{fig:key}, the first step of LATC is to convert the multivariate time series matrix $\boldsymbol{Y}$ into a tensor. We define an operator $\mathcal{Q}(\cdot)$, which converts the multivariate time series matrix into a third-order tensor. For instance, a partially observed matrix $\boldsymbol{Y}\in\mathbb{R}^{M\times (IJ)}$ can be converted into tensor $\mathcal{Q}(\boldsymbol{Y})\in\mathbb{R}^{M\times I\times J}$. Note that, given the size constraint, not all values are in $\boldsymbol{Y}$ to construct the tensor (see Figure~\ref{fig:key}). Correspondingly, $\mathcal{Q}^{-1}(\cdot)$ denotes the inverse operator that converts the third-order tensor into a multivariate time series matrix.

We define LATC as the following optimization problem:
\begin{equation}
    \begin{aligned}
    \min _{\boldsymbol{\mathcal{X}},\boldsymbol{Z},\boldsymbol{A}}~&\|\boldsymbol{\mathcal{X}}\|_{*}+\lambda\|\boldsymbol{Z}\|_{\boldsymbol{A},\mathcal{H}} \\ \text { s.t.}~&\left\{\begin{array}{l}\boldsymbol{\mathcal{X}}=\mathcal{Q}\left(\boldsymbol{Z}\right), \\ \mathcal{P}_{\Omega}(\boldsymbol{Z})=\mathcal{P}_{\Omega}(\boldsymbol{Y}), \\ \end{array}\right. \\
    \end{aligned}
    \label{lrtc_ar}
\end{equation}
where $\boldsymbol{Y}\in\mathbb{R}^{M\times (IJ)}$ is the partially observed time series matrix and $\boldsymbol{Z}$ has the same size as $\boldsymbol{Y}$, and $\lambda$ is a weight parameter that controls the trade-off between the two terms in the objective function. We define the \textbf{autoregressive norm} of matrix $\boldsymbol{Z}$ with a lag set $\mathcal{H}$ and coefficient matrix $\boldsymbol{A}$ as:
\begin{equation}
    \|\boldsymbol{Z}\|_{\boldsymbol{A},\mathcal{H}}=\sum_{m,t}(z_{m,t}-\sum_{i}a_{m,i}z_{m,t-h_i})^{2}.
    \label{auto_norm}
\end{equation}
Note that, with this definition, $\boldsymbol{A}$ is also a variable to estimate. For simplicity, we use independent autoregressive models in Eq.~\eqref{auto_norm} instead of a full vector autoregressive model. To solve the optimization problem, we perform the following transformation by introducing auxiliary variables $\boldsymbol{\mathcal{X}}_{k}$:
\begin{equation}
    \begin{aligned}
    \min _{\{\boldsymbol{\mathcal{X}}_{k}\}_{k=1}^{3},\boldsymbol{Z},\boldsymbol{A}}~&\sum_{k}\alpha_k\|\boldsymbol{\mathcal{X}}_{k(k)}\|_{*}+\lambda\|\boldsymbol{Z}\|_{\boldsymbol{A},\mathcal{H}} \\ \text { s.t.}~&\left\{\begin{array}{l}\boldsymbol{\mathcal{X}}_{k}=\mathcal{Q}\left(\boldsymbol{Z}\right),~k=1,2,3, \\ \mathcal{P}_{\Omega}(\boldsymbol{Z})=\mathcal{P}_{\Omega}(\boldsymbol{Y}). \\ \end{array}\right. \\
    \end{aligned}
    \label{lrtc_ar_admm}
\end{equation}

%In this formulation, we assume that: 1) The independent third-order tensors $\boldsymbol{\mathcal{X}}_{k}$s are low-rank in their mode-$k$ unfoldings, and these tensors can be converted into matrices by using the operator $\mathcal{Q}^{-1}(\cdot)$; 2) The matrix $\boldsymbol{Z}$ is an auxiliary variable which transforms observation information from $\boldsymbol{Y}$ to $\boldsymbol{\mathcal{X}}_{k}$s; 3) The variable $\boldsymbol{Z}$ follows an autoregressive process.

%For the fact that $\mathcal{Q}^{-1}(\boldsymbol{\mathcal{X}}_{k})=\boldsymbol{Z}$ as shown in Eq.~\eqref{lrtc_ar}, we consider a special case that solving the term of autoregressive process by reformulating this between two different variables. In what follows, we introduce Proposition~\ref{proposition1} for time series regression between two variables.

We next show the new optimization problem \eqref{lrtc_ar_admm} can be efficiently solved by employing the Alternating Direction Method of Multipliers (ADMM) framework. First, we introduce the following lemma, which allows us to estimate $\boldsymbol{A}$.
\begin{lemma}
\label{proposition1}
For two time series matrices $\boldsymbol{Z}=\boldsymbol{X}$ consisting of $M$ time series with $N$ time points, suppose the two variables follow an autoregressive model with a lag set $\mathcal{H}=\{h_1,\ldots,h_d\}$:
\begin{equation}\notag
    {z}_{m,t}=\sum_{i}{a}_{m,i}{x}_{m,t-h_i}+{\epsilon}_{m,t},\forall~m,t,
    \label{ar_regularizer}
\end{equation}
with $\epsilon_{mt}$ being Gaussian noise. Let $\boldsymbol{Q}_{m}=(\boldsymbol{v}_{h_d+1},\cdots,\boldsymbol{v}_{N})^\top\in\mathbb{R}^{(N-h_d)\times d}$ with $\boldsymbol{v}_{t}=({x}_{m,t-h_1},\cdots,{x}_{m,t-h_d})^\top$. Then, the solution to the problem
\begin{equation} \notag
\min_{\boldsymbol{a}_{m}}~\frac{1}{2}\|\boldsymbol{z}_{m,[h_d+1:]}-\boldsymbol{Q}_{m}\boldsymbol{a}_{m}\|_{2}^{2}
\end{equation}
can be written as
\begin{equation}
    \hat{\boldsymbol{a}}_{m}:=(\boldsymbol{Q}_{m}^\top\boldsymbol{Q}_{m})^{-1}\boldsymbol{Q}_{m}^\top\boldsymbol{z}_{m,[h_d+1:]}=\boldsymbol{Q}_{m}^{+}\boldsymbol{z}_{m,[h_d+1:]},
\end{equation}
where $\cdot^{+}$ denotes the pseudo-inverse of matrix.
\end{lemma}

% \begin{proof}
% It is easy to check that this optimization problem is equivalent to linear regression.
% \end{proof}

%In practice, applying Proposition~\ref{proposition1} to the ADMM solver for Eq.~\eqref{lrtc_ar} can guarantee the efficient parallel computing for the variable $\boldsymbol{Z}$. In this situation, we get the ADMM for the involved variables as follows,

Therefore, we can write down the following subproblems for ADMM:
\begin{align}
    \boldsymbol{\mathcal{X}}_{k}^{l+1}&:=\operatorname{arg}\min_{\boldsymbol{\mathcal{X}}}~\alpha_k\|\boldsymbol{\mathcal{X}}_{(k)}\|_{*}+\frac{\rho}{2}\|\mathcal{Q}^{-1}(\boldsymbol{\mathcal{X}})-\boldsymbol{Z}^{l}\|_{F}^{2}+\big\langle\mathcal{Q}^{-1}(\boldsymbol{\mathcal{X}})-\boldsymbol{Z}^{l},\mathcal{Q}^{-1}(\boldsymbol{\mathcal{T}}_{k}^{l})\big\rangle,\label{tensor_sub1} \\
    \boldsymbol{Z}^{l+1}_{[:h_d]}&:=
    \operatorname{arg}\min_{\boldsymbol{Z}}~\sum_{k}\left(\frac{\rho}{2}\|\mathcal{Q}^{-1}(\boldsymbol{\mathcal{X}}_{k}^{l+1})_{[:h_d]}-\boldsymbol{Z}\|_{F}^{2}
    +\big\langle\mathcal{Q}^{-1}(\boldsymbol{\mathcal{X}}_{k}^{l+1})_{[:h_d]}-\boldsymbol{Z},\mathcal{Q}^{-1}(\boldsymbol{\mathcal{T}}_{k}^{l})_{[:h_d]}\big\rangle\right),  \label{tensor_sub2} \\
    \boldsymbol{Z}^{l+1}_{[h_d+1:]}&
    \begin{aligned}[t]
    &:=\operatorname{arg}\min_{\boldsymbol{Z}}~\sum_{m}\frac{\lambda}{2}\|\boldsymbol{z}_{m,[h_d+1:]}-\boldsymbol{Q}_{m}\boldsymbol{a}_{m}\|_{2}^{2}+\\ &\sum_{k}\left(\frac{\rho}{2}\|\mathcal{Q}^{-1}(\boldsymbol{\mathcal{X}}_{k}^{l+1})_{[h_d+1:]}-\boldsymbol{Z}\|_{F}^{2}+\big\langle\mathcal{Q}^{-1}(\boldsymbol{\mathcal{X}}_{k}^{l+1})_{[h_d+1:]}-\boldsymbol{Z},\mathcal{Q}^{-1}(\boldsymbol{\mathcal{T}}_{k}^{l})_{[h_d+1:]}\big\rangle\right)\\
    \end{aligned} \label{tensor_sub3} \\
    \boldsymbol{a}_{m}^{l+1}&:=\boldsymbol{Q}_{m}^{+}\boldsymbol{z}^{l+1}_{m,[h_d+1:]}, \label{tensor_sub5} \\
    \boldsymbol{\mathcal{T}}_{k}^{l+1}&:=\boldsymbol{\mathcal{T}}_{k}^{l}+\rho(\boldsymbol{\mathcal{X}}_{k}^{l+1}-\mathcal{Q}(\boldsymbol{Z}^{l+1})), \label{tensor_sub4}
\end{align}
where the symbol $ \left<\cdot,\cdot\right>$ denotes the inner product, $\rho$ is the learning rate of ADMM algorithm, and $l$ denotes the count of iteration. The matrix $\boldsymbol{Q}_{m}$ used in Eq.~\eqref{tensor_sub3} and \eqref{tensor_sub5} are defined on the matrix $\mathcal{Q}^{-1}(\hat{\boldsymbol{\mathcal{X}}})$ where $\hat{\boldsymbol{\mathcal{X}}}$ is the estimated tensor at iteration $l+1$: $\hat{\boldsymbol{\mathcal{X}}}=\sum_{k}\alpha_k\boldsymbol{\mathcal{X}}_{k}^{l+1}$.

The subproblem \eqref{tensor_sub1} for computing $\boldsymbol{\mathcal{X}}_k$ is convex, and the closed-form solution is given by
\begin{equation}
    \boldsymbol{\mathcal{X}}_{k}^{l+1}:=\operatorname{fold}_{k}\left(\mathcal{D}_{\alpha_k/\rho}(\mathcal{Q}(\boldsymbol{Z}^{l})_{(k)}-\boldsymbol{\mathcal{T}}_{k(k)}^{l}/\rho)\right),\label{solution1}
\end{equation}
where the symbol $\mathcal{D}_{\alpha/\rho}(\cdot)$ denotes the operator of singular value thresholding with shrinkage parameter $\alpha/\rho$. The solution in Eq.~\eqref{solution1} meets the singular value thresholding as shown in Lemma~\ref{lemma1}.

\begin{lemma}
\label{lemma1}
For any $\alpha,\rho>0$, and $\boldsymbol{Z}\in\mathbb{R}^{M\times N}$, a global optimal solution to the problem $\min_{\boldsymbol{X}}~\alpha\|\boldsymbol{X}\|_{*}+\frac{1}{2}\rho\|\boldsymbol{X}-\boldsymbol{Z}\|_{F}^{2}$ is given by the singular value thresholding \cite{cai2010svt}:
\begin{equation}
    \hat{\boldsymbol{X}}:=\mathcal{D}_{\alpha/\rho}(\boldsymbol{Z})=\boldsymbol{U}\diag\left([\boldsymbol{\sigma}(\boldsymbol{Z})-\alpha/\rho]_{+}\right)\boldsymbol{V}^\top,
\end{equation}
where $\boldsymbol{Z}=\boldsymbol{U}\diag(\boldsymbol{\sigma}(\boldsymbol{Z}))\boldsymbol{V}^\top$ is the singular value decomposition of $\boldsymbol{Z}$. The symbol $[\cdot]_{+}$ denotes the positive truncation at 0 which satisfies $[\boldsymbol{\sigma}-\alpha/\rho]_{+}=\max\{\boldsymbol{\sigma}-\alpha/\rho,\boldsymbol{0}\}$.
\end{lemma}

% \begin{proof}
% This proof can be found in \cite{x}.
% \end{proof}

For variable $\boldsymbol{Z}$, the subproblems~\eqref{tensor_sub2} and ~\eqref{tensor_sub3} are both convex least squares. We can therefore derive the closed-form solution
\begin{align}
    \boldsymbol{Z}_{[:h_d]}^{l+1}:=&\frac{1}{3}\sum_k\mathcal{Q}^{-1}(\boldsymbol{\mathcal{X}}_{k}^{l+1}+\boldsymbol{\mathcal{T}}_{k}^{l}/\rho)_{[:h_d]},\label{solution2} \\
    \boldsymbol{z}_{m,[h_d+1:]}^{l+1}:=&\frac{1}{3(\rho+\lambda)}\sum_{k}\mathcal{Q}^{-1}(\rho\boldsymbol{\mathcal{X}}_{k}^{l+1}+\boldsymbol{\mathcal{T}}_{k}^{l})_{m,[h_d+1:]}+\frac{\lambda}{\rho+\lambda}\boldsymbol{Q}_{m}\boldsymbol{a}_{m}^{l},\label{solution3}
\end{align}
where we impose a fixed consistency constraint, namely $\mathcal{P}_{\Omega}(\boldsymbol{Z}^{l+1}):=\mathcal{P}_{\Omega}(\boldsymbol{Y})$, to guarantee the transformation of observation information at each iteration. In addition, for Eq.~\eqref{solution3}, we can set $\lambda$ as $c_0\cdot\rho$ with $c_0$ being a constant determining the relative weight of time series regression.

Until now, we use the NN in the objective function of LATC. In fact, another way to capture global low-rank patterns is through the TNN minimization, which is experimentally proved to be better than NN minimization \cite{hu2013fast}. In this paper, we also test a variant of LATC based on TNN minimization, and solve the following subproblem for updating $\boldsymbol{\mathcal{X}}_{k}$:
\begin{equation}
    \boldsymbol{\mathcal{X}}_{k}^{l+1}:=\operatorname{arg}\min_{\boldsymbol{\mathcal{X}}}~\alpha_k\|\boldsymbol{\mathcal{X}}_{(k)}\|_{\theta,*}+\frac{1}{2}\rho\|\mathcal{Q}^{-1}(\boldsymbol{\mathcal{X}})-\boldsymbol{Z}^{l}\|_{F}^{2}+\big\langle\mathcal{Q}^{-1}(\boldsymbol{\mathcal{X}})-\boldsymbol{Z}^{l},\mathcal{Q}^{-1}(\boldsymbol{\mathcal{T}}_{k}^{l})\big\rangle,
\label{tnn_sub1}
\end{equation}
where $\theta$ is a nonnegative integer. The TNN minimization reduces to NN minimization when $\theta=0$. Thus, Eq.~\eqref{tensor_sub1} is indeed a special case of Eq.~\eqref{tnn_sub1}. If we integrate Lemma~\ref{lemma2} into Eq.~\eqref{tnn_sub1}, we get
\begin{equation}
    \boldsymbol{\mathcal{X}}_{k}^{l+1}:=\operatorname{fold}_{k}\left(\mathcal{D}_{\theta,\alpha_k/\rho}(\mathcal{Q}(\boldsymbol{Z}^{l})_{(k)}-\boldsymbol{\mathcal{T}}_{k(k)}^{l}/\rho)\right).\label{tnn_solution1}
\end{equation}

\begin{lemma}
\label{lemma2}
For any $\alpha,\rho>0$, $\boldsymbol{Z}\in\mathbb{R}^{M\times N}$, and $\theta\in\mathbb{N}_{+}$ where $\theta<\min\{M,N\}$, an optimal solution to the problem $\min_{\boldsymbol{X}}~\alpha\|\boldsymbol{X}\|_{\theta,*}+\frac{1}{2}\rho\|\boldsymbol{X}-\boldsymbol{Z}\|_{F}^{2}$ is given by the generalized singular value thresholding \cite{zhang2011penalty,chen2013reduced,lu2015generalized}:
\begin{equation}
    \hat{\boldsymbol{X}}:=\mathcal{D}_{\theta,\alpha/\rho}(\boldsymbol{Z})=\boldsymbol{U}\diag\left([\boldsymbol{\sigma}(\boldsymbol{Z})-\mathfrak{1}_{\theta}\cdot\alpha/\rho]_{+}\right)\boldsymbol{V}^\top,\label{tnn_solution}
\end{equation}
where $\boldsymbol{Z}=\boldsymbol{U}\diag(\boldsymbol{\sigma}(\boldsymbol{Z}))\boldsymbol{V}^\top$ is the singular value decomposition of $\boldsymbol{Z}$. The symbol $[\cdot]_{+}$ denotes the positive truncation at 0 as defined in Lemma~\ref{lemma1}. $\mathfrak{1}_{\theta}\in\{0,1\}^{\min\{M,N\}}$ is a binary indicator vector whose first $\theta$ entries are 0 and other entries are 1.
\end{lemma}

\section{Experiments}

\subsection{Experiment setup}
In this section, we assess the performance of LATC using three real-world multivariate time series data sets: (1) \textbf{PeMS}\footnote{from \url{http://pems.dot.ca.gov/} and \url{https://github.com/VeritasYin/STGCN_IJCAI-18}} (P) registers traffic speed time series from 228 sensors over 44 days with 288 time points per day (i.e., 5-min frequency). (2) \textbf{Guangzhou}\footnote{from \url{https://doi.org/10.5281/zenodo.1205228}} (G) contains traffic speed time series from 214 road segments in Guangzhou, China over 61 days with 144 time points per day (i.e., 10-min frequency). (3) \textbf{Electricity}\footnote{from \url{https://archive.ics.uci.edu/ml/datasets/ElectricityLoadDiagrams20112014}} (E) records hourly electricity consumption transactions of 370 clients from 2011 to 2014. We use a subset of the last five weeks of 321 clients in our experiments.

Our experiments cover the same two tasks as in \cite{yu2016temporal}: missing data \textit{imputation} and rolling \textit{prediction}. As mentioned, in LATC, both imputation and prediction are achieved by performing tensor completion. For missing data imputation, we follow the default LATC framework as a general \textit{imputer} procedure (see Algorithm~\ref{imputer}). The default prediction task follows the description in Figure~\ref{fig:key}, in which we use the recovered values of the red submatrix as the prediction for the future $\tau$
time points as a window. Rolling prediction for multiple windows is obtained by applying LATC repeatedly. Algorithm~\ref{predictor} summarizes the rolling \textit{predictor} procedure for $S$ rolling windows.

The sizes $M\times I\times J$ of the transformed tensors of the three data sets are: 228$\times$288$\times$44 for (P), 214$\times$144$\times$61 for (G), and 321$\times$24$\times$35 for (E). For the imputation task, we randomly mask certain amount (20\%/40\%) of values as missing. We consider two missing scenarios: random missing (RM) in which entries are missing randomly, and non-random missing (NM) in which each time series has block missing for randomly selected days (i.e., randomly removing mode-3 fibers in tensor $\mathcal{Q}(\boldsymbol{Y})$). For data set (P), we perform rolling predictions for the last 5 days (i.e., 1440 time points) with $\tau=9$ and $S=160$ for in the presence of missing values. Similarly, we also predict the last 7 days for data set (G) with $\tau=12$ and $S=84$ and the last 5 days for data set (E) with $\tau=6$ and $S=20$. We use $\text{MAPE}=\frac{1}{n}\sum_{i=1}^n |\frac{x_i-\hat{x}_i}{x_i}|\times 100$ and $\text{RMSE}=\sqrt{\frac{1}{n}\sum_{i=1}^n \left(x_i-\hat{x}_i\right)^2}$ as performance metrics. The adapted data sets and Python implementation for these experiments are available in our GitHub repository \url{https://github.com/xinychen/tensor-learning}.

\begin{algorithm}[!ht]
\caption{$\text{imputer}(\boldsymbol{Y},\boldsymbol{\alpha},\rho,\lambda,\theta)$}
\label{imputer}
Initialize $\boldsymbol{\mathcal{T}}$ as zeros and $\boldsymbol{A}$ as small random values. Set $\mathcal{P}_{\Omega}(\boldsymbol{Z})=\mathcal{P}_{\Omega}(\boldsymbol{Y})$ and $l=0$. \\
\While{not converged}{
\For{$k = 1$ \KwTo $3$}{
    Update $\boldsymbol{\mathcal{X}}_{k}^{l+1}$ by Eq.~\eqref{solution1} for NN or Eq.~\eqref{tnn_solution1} for TNN;
    }

Compute $\hat{\boldsymbol{\mathcal{X}}}=\sum_k\alpha_k\boldsymbol{\mathcal{X}}_{k}^{l+1}$, and then $\hat{\boldsymbol{X}}=\mathcal{Q}^{-1}(\hat{\boldsymbol{\mathcal{X}}})$; \\
Update $\boldsymbol{Z}_{[:h_d]}^{l+1}$ by Eq.~\eqref{solution2}; \\
\For{$m=1$ \KwTo $M$}{
    Build $\boldsymbol{Q}_{m}$ on $\hat{\boldsymbol{X}}$, and then update $\boldsymbol{z}_{m,[h_d+1:]}^{l+1}$ by Eq.~\eqref{solution3};
}

Transform observation information by letting  $\mathcal{P}_{\Omega}(\boldsymbol{Z}^{l+1})=\mathcal{P}_{\Omega}(\boldsymbol{Y})$; \\
\For{$m=1$ \KwTo  $M$}{
    Update $\boldsymbol{a}_{m}^{l+1}$ by Eq.~\eqref{tensor_sub5};
}

Update $\boldsymbol{\mathcal{T}}^{l+1}$ by Eq.~\eqref{tensor_sub4};\\
$l:=l+1$;\\
$\rho=\min\{1.05\times\rho, \rho_{\text{max}}\}$;
}

\Return recovered matrix $\hat{\boldsymbol{X}}$.
\end{algorithm}

\begin{algorithm}[!ht]
\caption{$\text{predictor}(\boldsymbol{Y},t,S,\tau,\boldsymbol{\alpha},\rho,\lambda,\theta)$\quad\# $S$ is the total number of rolling windows}
\label{predictor}
Initialize $M$-by-$S\tau$ matrix $\tilde{\boldsymbol{X}}$ with zeros. \\
\For{$s=1$ \KwTo $S$}{
    Stacking data $\boldsymbol{Y}$ from time point $t+s\tau-IJ$ to $t+s\tau$ as $\boldsymbol{Y}_{s}$, in which the last $\tau$ columns (to be predicted) are masked as missing values; \\
    $\hat{\boldsymbol{X}}_{s}=\text{imputer}(\boldsymbol{Y}_{s},\boldsymbol{\alpha},\rho,\lambda,\theta)$; \\
    $\tilde{\boldsymbol{X}}_{[s(\tau-1)+1:s\tau]}:=\hat{\boldsymbol{X}}_{s[IJ-\tau+1:]}$;
}

\Return predicted matrix $\tilde{\boldsymbol{X}}$.
\end{algorithm}

\subsection{Baseline models}
We compare LATC with some state-of-the-art approaches, including: (1) Temporal Regularized Matrix Factorization (TRMF) \cite{yu2016temporal}, which is an autoregression regularized temporal matrix factorization. (2) Bayesian Temporal Matrix Factorization (BTMF) \cite{sun2019bayesian}, which is a fully Bayesian matrix factorization model by integrating vector autoregressive process into the latent temporal factors. (3) High-accuracy Low-Rank Tensor Completion (HaLRTC) \cite{liu2013tensor}, which minimizes NN to achieve completion. (4) HaLRTC-TNN, which replaces NN with TNN in objective function. For the LATC framework, we build two variants: LATC-NN with NN minimization and LATC-TNN with TNN minimization. The detailed settings for these models are presented in Appendix~\ref{app:parameter}.

% {\color{red}For each missing scenario, we randomly select 5\% for cross validation to tune parameters for these models. The detailed parameter settings are give in Appendix~\ref{app:parameter}.}

\subsection{Results}

\begin{table}[!ht]
  \caption{Imputation performance (MAPE/RMSE).}
\small
  \label{imputation_results}
  \centering
  \begin{tabular}{crrrrrrr}
    \toprule
    Models & TRMF & BTMF & HaLRTC & HaLRTC-TNN & LATC-NN & LATC-TNN \\
    \midrule
    20\%, RM (P) & 5.68/3.87 & 5.82/3.96 & 5.92/3.90 & 5.21/3.60 & 3.36/2.32 & \textbf{2.97}/\textbf{2.14} \\
    40\%, RM (P) & 5.75/3.92 & 5.93/4.02 & 7.05/4.56 & 6.08/4.18 & 4.13/2.84 & \textbf{3.50}/\textbf{2.54} \\
    20\%, NM (P) & 9.41/6.27 & 9.40/6.26 & 8.72/5.64 & 7.81/5.36 & 8.79/5.65 & \textbf{7.31}/\textbf{5.15} \\
    40\%, NM (P) & 9.54/6.40 & 9.51/6.39 & 9.46/6.04 & 8.33/5.69 & 9.70/6.12 & \textbf{7.78}/\textbf{5.46} \\
    \midrule
    20\%, RM (G) & 7.25/3.11 & 7.39/3.15 & 8.14/3.33 & 6.73/2.88 & {7.12}/{2.97} & \textbf{6.28}/\textbf{2.73} \\
    40\%, RM (G) & 7.40/3.19 & 7.63/3.27 & 8.87/3.61 & 7.27/3.12 & {7.82}/{3.24} & \textbf{6.79}/\textbf{2.96} \\
    20\%, NM (G) & 10.19/4.28 & 10.17/4.27 & 10.46/4.21 & \textbf{9.32}/3.96 & 10.46/{4.21} & 9.33/\textbf{3.95} \\
    40\%, NM (G) & 10.37/4.46 & 10.38/4.48 & 10.88/4.38 & \textbf{9.51}/4.08 & 10.89/4.38 & \textbf{9.51}/\textbf{4.07} \\
    \midrule
    20\%, RM (E) & 13.12/723 & 12.85/948 & 10.36/530 & 10.20/\textbf{482} & 9.79/527 & \textbf{9.71}/530 \\
    40\%, RM (E) & 13.63/862 & 13.34/1281 & 11.30/689 & 11.15/571 & 10.66/\textbf{738} & \textbf{10.59}/789 \\
    20\%, NM (E) & 26.31/3665 & 19.72/1623 & 16.93/2260 & 16.83/728 & \textbf{16.55}/802 & 16.58/\textbf{652} \\
    40\%, NM (E) & 22.71/2941 & 18.00/1817 & 15.86/4921 & 15.70/1769 & 15.51/1467 & \textbf{15.50}/\textbf{1026} \\
    % 40\%, RM (E) & 37.34/0.005 & 23.60/0.008 & 20.92/0.011 & 21.48/0.033 & 15.60/\textbf{0.004} & \textbf{15.40}/0.007 \\
    % 20\%, NM (E) & 49.96/\textbf{0.008} & 29.44/0.021 & 28.44/0.031 & 30.07/0.044 & \textbf{28.26}/0.009 & 28.51/0.011 \\
    % 40\%, NM (E) & 51.08/\textbf{0.012} & 38.55/0.017 & \textbf{32.19}/0.037 & 32.26/0.043 & 33.79/0.014 & 32.79/0.016 \\
    \bottomrule
  \end{tabular}
\end{table}

\textbf{Imputation Results } Table~\ref{imputation_results} shows the results for imputation tasks. As can be seen, the proposed LATC achieves the best imputation accuracy in almost all cases. Essentially, TNN-based models offer better performance than NN-based models. The superiority of LATC over HaLRTC clearly shows that the autoregressive norm can better capture temporal dynamics than the pure low-rank structure. On the other hand, LATC also outperforms the two matrix-based models: TRMF and BTMF. The result suggests that LATC can effectively leverage the global (i.e., ``daily'' for all three data sets) patterns and consistency on the temporal dimension, which is difficult to model using the local autoregressive dynamics alone in the matrix representation.

%In addition, the tensor representation can deal with certain extreme missing scenarios such as ``blackouts''---missing of several consecutive columns in the original matrix.

\textbf{Prediction Results } Table~\ref{forecast_results_2} shows the results for rolling prediction tasks. We use the same set of autoregressive lags for TRMF, BTMF, and LATC. As can be seen, the proposed LATC outperforms other models by a substantial margin. Although TRMF an BTMF are powerful in capturing the global consistency/similarity among sensors, the AR model alone is insufficient in capture the temporal patterns at different scales. Moreover, TRMF and BTMF work on the latent layer, which may ignore the local property of each time series. In this case, the tensor representation shows clear advantage. Similar to the imputation task, we find that LATC-TNN essentially gives better results than its NN-based counterpart in most cases. By comparing HaLRTC and LATC-NN side by side, we can clearly see the importance of the autoregressive norm in LATC. Appendix~\ref{imputation-prediction-example} provides some example results for both the imputation and prediction tasks as figures.

% When producing forecasts without missing values, Table~\ref{forecast_results_1} shows that LATC's forecast RMSE is lower than STGCN and DeepGLO while our MAPE is slightly higher than STGCN. This indicates that the performance of our model is still competitive even in the case of no-missing data. In presence of missing values, these two deep learning models---STGCN and DeepGLO---cannot produce accurate forecasts anymore. Our experimental results in Table~\ref{forecast_results_2} demonstrate that LATC's forecast MAPE/RMSE continues to be lower than the baseline models on PeMS data. On Guangzhou data, we find that LATC performs significantly better than the baseline models while LATC still produces competitive forecasts in presence of NM data. Fig.~\ref{PeMS_forecasting} shows some forecasting examples reported by LATC in presence of RM data.

%In addition, we compete our model against STGCN \cite{yu2018stgcn} and DeepGLO \cite{sen2019think} for time series forecasting without missing values.

% \begin{table}
% \small
%   \caption{Time series forecasting performance.}
%   \label{forecast_results_1}
%   \centering
%   \begin{tabular}{l|ccccc}
%     \toprule
%      & HA & STGCN & DeepGLO & LATC-TNN \\
%     \midrule
%     Original (P) & 10.61/7.20 & 6.24/5.27 & 7.90/6.49 & 6.50/4.96 \\
%     Original (G) & 13.48/5.14 & -/- & -/- & 11.37/4.55 \\
%     \bottomrule
%   \end{tabular}
% \end{table}

\begin{table}
  \caption{Rolling prediction performance in the presence of missing values (MAPE/RMSE).}
\small
  \label{forecast_results_2}
  \centering
  \begin{tabular}{crrrrrrr}
    \toprule
    Models & TRMF & BTMF & HaLRTC & HaLRTC-TNN & LATC-NN & LATC-TNN \\
    \midrule
    Original (P) & 11.30/7.19 & 8.83/5.95 & 9.98/6.21 & 7.51/5.37 & 6.62/4.99 & \textbf{6.39}/\textbf{4.97} \\
    20\%, RM (P) & 10.57/6.79 & 8.84/5.99 & 10.09/6.27 & 7.64/5.44 & 6.77/5.08 & \textbf{6.53}/\textbf{5.07} \\
    40\%, RM (P) & 10.26/6.64 & 8.97/6.03 & 10.27/6.37 & 7.81/5.54 & 6.99/5.21 & \textbf{6.82}/\textbf{5.16} \\
    20\%, NM (P) & 11.21/7.07 & 9.02/6.03 & 10.35/6.39 & 7.73/5.50 & 7.96/5.50 & \textbf{7.32}/\textbf{5.35} \\
    40\%, NM (P) & 11.90/7.31 & 9.47/6.41 & 10.90/6.68 & 8.09/5.73 & 9.18/6.09 & \textbf{7.88}/\textbf{5.65} \\
    \midrule
    Original (G) & 13.33/5.22 & 11.38/4.64 & 12.79/4.88 & \textbf{10.39}/\textbf{4.29} & 11.11/4.52 & \textbf{10.39}/\textbf{4.29} \\
    20\%, RM (G) & 13.34/5.22 & 11.49/4.67 & 12.86/4.90 & 10.43/\textbf{4.30} & 11.24/4.54 & \textbf{10.42}/\textbf{4.30} \\
    40\%, RM (G) & 13.46/5.19 & 11.54/4.70 & 12.98/4.94 & \textbf{10.47}/\textbf{4.32} & 11.44/4.59 & 10.48/4.33 \\
    20\%, NM (G) & 13.84/5.30 & 11.62/4.74 & 13.05/4.96 & \textbf{10.47}/\textbf{4.33} & 12.02/4.68 & 10.48/4.34 \\
    40\%, NM (G) & 14.58/5.55 & 11.74/4.80 & 13.47/5.10 & \textbf{10.67}/\textbf{4.42} & 12.67/4.87 & \textbf{10.67}/\textbf{4.42} \\
    \midrule
    Original (E) & 28.37/1154 & 27.83/1016 & 25.48/953 & \textbf{24.94}/\textbf{779} & 25.48/953 & \textbf{24.94}/\textbf{779} \\
    20\%, RM (E) & 27.88/1130 & 28.20/1023 & 25.87/983 & \textbf{26.31}/\textbf{863} & 25.87/983 & \textbf{26.31}/\textbf{863} \\
    40\%, RM (E) & 28.64/1336 & 28.50/1209 & 26.58/1042 & 26.63/\textbf{890} & \textbf{26.07}/981 & 26.63/\textbf{890} \\
    20\%, NM (E) & 28.99/1142 & 31.07/1335 & 27.67/1536 & \textbf{25.15}/\textbf{811} & 27.67/1536 & 26.78/861 \\
    40\%, NM (E) & 28.68/1472 & 32.46/1718 & 26.92/2179 & 27.19/899 & \textbf{24.98}/1271 & 27.00/\textbf{888} \\
    % \midrule
    % Original (E) & & & 74.85/0.006 & & 47.00/0.004 & 44.74/0.005 \\
    % 20\%, RM (E) & & & 77.77/0.008 & & 50.46/0.004 & 47.59/0.005 \\
    % 40\%, RM (E) & & & 76.20/0.015 & & 51.36/0.005 & 46.51/0.005 \\
    % 20\%, NM (E) & & & 71.39/0.011 & & 50.23/0.005 & 50.58/0.006 \\
    % 40\%, NM (E) & & & 72.11/0.018 & & 53.71/0.006 & 54.79/0.007 \\
    \bottomrule
  \end{tabular}
\end{table}

\section{Conclusion}

We proposed LATC as a new framework to model large-scale multivariate time series data with missing values. By transforming the original matrix to a tensor, LATC can model both imputation and prediction as a universal tensor completion problem in which all observed data will contribute to the final prediction. We impose low-rank assumption to capture global patterns across all the three dimensions (sensor, time of day, and day), and further introduce a novel autoregressive norm to characterize local temporal trends.  Our numerical experiment on three real-world data sets further confirms the importance of incorporating both global patterns and local trends in time series models. This study can be extended in several ways. A major limitation of LATC is its high computational cost: we have to train a new model for each prediction window. It will be interesting to develop strategies to avoid re-training, and making the prediction model online. LATC can also be extended to a high-dimensional setting for matrix and tensor time series data \cite{sun2019bayesian,jing2018high}. In addition, if side information on sensors are available (e.g., location and network structure), additional regularizers can be introduced to impose local consistency for sensors \cite{bahadori2014fast}.

\section*{Acknowledgement}
This research is supported by the Natural Sciences and Engineering Research Council (NSERC) of Canada, the Institute for Data Valorisation (IVADO), and the Canada Foundation for Innovation (CFI). We would like to thank Jinming Yang from Shanghai Jiao Tong University for helpful discussion.

%\section*{Broader impact}
%This research provides a new perspective to model the multivariate time series prediction problem. The model is particularly designed for common real-world scenarios in which one would like to perform accurate prediction in the presence of data corruption and missing values. This is a universal problem to many science and engineering domains, and LATC provides a new method to solve the problem. We show the importance of both global and local patterns in time series data, and hope our work can help design better  sequence-learning models in deep learning.

% Our work aims to effectively acquire knowledge from such datasets to support decision making processes in various domains such as climatology, smart city, public health, and ecology. The open-source code of this paper will be disseminated through the researches who are interested in this topic.

% This research presents a novel low-rank autoregressive framework which is well-suited for modeling multivariate time series data with low-rank property. This framework has great potential to contribute to the time series analysis problems (e.g., multivariate time series imputation/prediction) that involve missing values. Moreover, this framework will provide an important alternative for developing time series models.

% \section*{References}

\bibliographystyle{IEEEtran}
\bibliography{var}

\newpage

\appendix

\begin{center}
    \Large\textbf{Supplementary Material}
\end{center}

\section{Parameter setting} \label{app:parameter}

In this section, we give the parameter setting for our experiments. Note that all experiments were tested using Python 3.7 on a laptop with 2.3 GHz Intel Core i5 (CPU) and 8 GB RAM.

\subsection{HaLRTC, HaLRTC-TNN, LATC-NN, and LATC-TNN}

In our experiments, given season length $I$, we set the time lags to  $\mathcal{H}=\left\{1,\ldots,6,I-2,\ldots,I+3\right\}$ for each data set. For instance, in Electricity data, we have season length $I=24$ and set the time lags as $\{1,2,...,6,22,23,...,27\}$. To determine the convergence of the algorithm, we use $$\mathcal{C}^{l+1}=\frac{\|\hat{\boldsymbol{X}}^{l+1}-\hat{\boldsymbol{X}}^{l}\|_{F}}{\|\mathcal{P}_{\Omega}({\boldsymbol{Y}})\|_{F}}<\epsilon$$ as convergence condition, where $\hat{\boldsymbol{X}}^{l+1}$ and $\hat{\boldsymbol{X}}^{l}$ denote the recovered matrices at the $l+1$th iteration and $l$th iteration, respectively. For reaching convergence, we set $\epsilon=0.0001$ for the algorithm.

% Table~\ref{parameter_LATC} shows the tuned parameter setting for LATC on both PeMS data, Guangzhou data, and Electricity data. Given season length $I$, we set the time lags to  $\mathcal{H}=\left\{1,\ldots,6,I-2,\ldots,I+3\right\}$ for each data set. We choose $\rho$ from $5\times 10^{-7},10^{-6},5\times 10^{-6},10^{-5},5\times 10^{-5},10^{-4},5\times 10^{-4},10^{-3}\}$, $\lambda$ from $\{0, 0.1\times  \rho, 0.5\times  \rho,1\times  \rho,5\times  \rho,10\times  \rho\}$, and $\theta$ from $\{1,3,5,10,15,30,50\}$ for validation. We use $\mathcal{C}^{l+1}=\frac{\|\hat{\boldsymbol{X}}^{l+1}-\hat{\boldsymbol{X}}^{l}\|_{F}}{\|\mathcal{P}_{\Omega}({\boldsymbol{Y}})\|_{F}}<\epsilon$ as convergence condition, where $\hat{\boldsymbol{X}}^{l+1}$ and $\hat{\boldsymbol{X}}^{l}$ denote the recovered matrices at the $l+1$th iteration and $l$th iteration, respectively. For reaching convergence, we set $\epsilon=0.0001$ for the algorithm.

For imputation tasks, we set parameters of LATC-TNN for each data set as:
\begin{itemize}
    \item (\textbf{P}) PeMS data: For RM scenarios, we set parameters $\rho=0.0001$, $\lambda=5\times\rho$, and $\theta=15$ for RM scenarios. For NM scenarios, we set $\rho=0.0001$, $\lambda=1\times\rho$, and $\theta=10$.
    \item (\textbf{G}) Guangzhou data: For RM scenarios, we set parameters $\rho=0.0001$, $\lambda=5\times\rho$, and $\theta=30$. For NM scenarios, we set $\rho=0.0001$, $\lambda=1\times\rho$, and $\theta=10$.
    \item (\textbf{E}) Electricity data: For RM scenarios, we set parameters $\rho=0.000001$, $\lambda=5\times\rho$, $\theta=5$ (for 20\% missing), and $\theta=3$ (for 40\% missing). For NM scenarios, we set $\rho=0.000001$, $\lambda=5\times\rho$, and $\theta=1$.
\end{itemize}
Here, HaLRTC is a special case of LATC-NN (i.e., with $\lambda=0$), and we evaluate the HaLRTC imputer/predictor with same $\rho$. Similarly, HaLRTC-TNN is a special case of LATC-TNN (i.e., with $\lambda=0$), and we also evaluate the HaLRTC-TNN imputer with same $\rho$ and $\theta$. To evaluate LATC-NN, we let $\theta=0$ in the parameters of LATC-TNN.

For prediction tasks, we choose parameters by testing on validation set. Table~\ref{predictor_parameter} shows the tuned parameter setting for HaLRTC, HaLRTC-TNN, LATC-NN, and LATC-TNN on all three data sets according to validation RMSEs. The parameter set and validation set for each data set are given as:
\begin{itemize}
    \item (\textbf{P}) PeMS data: We choose $\rho$ from $\{0.0001, 0.0005,0.001\}$, $\lambda$ from $\rho\times\{0,0.1,0.5,1,5,10\}$, and $\theta$ from $\{0,5,10,15\}$ by predicting the last 20-window time series (i.e., validation set) before the last 5 days (i.e., testing set).
    \item (\textbf{G}) Guangzhou data: We choose $\rho$ from $\{0.0001,0.0005,0.001\}$, $\lambda$ from $\rho\times\{0,0.1,0.5,1,5,10\}$, and $\theta$ from $\{0,5,10,15\}$ by predicting the last 10-window time series (i.e., validation set) before the last 7 days (i.e., testing set).
    \item (\textbf{E}) Electricity data: We choose $\rho$ from $\{0.0000001,0.000001,0.00001\}$, $\lambda$ from $\rho\times\{0,0.1,0.5,1,5,10\}$, and $\theta$ from $\{1,3,5,10,15\}$ by predicting the last 10-window time series (i.e., validation set) before the last 5 days (i.e., testing set).
\end{itemize}

\begin{table}[!ht]
\small
  \caption{Parameter setting for HaLRTC, HaLRTC-TNN, LATC-NN, and LATC-TNN predictors on real-world data sets where $\epsilon=0.0001$ is the stop criterion for convergence.}
  \label{predictor_parameter}
  \centering
  \begin{tabular}{l|c|cc|cc|ccc}
    \toprule
     & HaLRTC & \multicolumn{2}{c|}{HaLRTC-TNN} & \multicolumn{2}{c|}{LATC-NN} & \multicolumn{3}{c}{LATC-TNN} \\
     & $\rho$ & $\rho$ & $\theta$ & $\rho$ & $\lambda$ & $\rho$ & $\lambda$ & $\theta$ \\
    \midrule
     Original (P) & 0.0001 & 0.0001 & 15 & 0.0005 & $10\times\rho$ & 0.0005 & $10\times\rho$ & 10 \\
     20\%, RM (P) & 0.0001 & 0.0001 & 15 & 0.001 & $5\times\rho$ & 0.0005 & $10\times\rho$ & 10 \\
     40\%, RM (P) & 0.0001 & 0.0001 & 15 & 0.0005 & $10\times\rho$ & 0.0005 & $5\times\rho$ & 15 \\
     20\%, NM (P) & 0.0001 & 0.0001 & 15 & 0.0005 & $5\times\rho$ & 0.0005 & $5\times\rho$ & 5 \\
     40\%, NM (P) & 0.0001 & 0.0001 & 15 & 0.0005 & $5\times\rho$ & 0.0001 & $5\times\rho$ & 15 \\
    \midrule
     Original (G) & 0.0001 & 0.0001 & 10 & 0.0005 & $5\times\rho$ & 0.0001 & $0.5\times\rho$ & 10 \\
     20\%, RM (G) & 0.0001 & 0.0001 & 10 & 0.0005 & $5\times\rho$ & 0.0001 & $0.1\times\rho$ & 10 \\
     40\%, RM (G) & 0.0005 & 0.0001 & 10 & 0.0005 & $5\times\rho$ & 0.0001 & $0.5\times\rho$ & 10 \\
     20\%, NM (G) & 0.0001 & 0.0001 & 15 & 0.0001 & $10\times\rho$ & 0.0001 & $0.1\times\rho$ & 15 \\
     40\%, NM (G) & 0.0001 & 0.0001 & 15 & 0.0001 & $10\times\rho$ & 0.0001 & $0\times\rho$ & 15 \\
    \midrule
     Original (E) & 0.000001 & 0.0000001 & 5 & 0.000001 & $0\times\rho$ & 0.0000001 & $0\times\rho$ & 5 \\
     20\%, RM (E) & 0.000001 & 0.0000001 & 1 & 0.000001 & $0\times\rho$ & 0.0000001 & $0\times\rho$ & 1 \\
     40\%, RM (E) & 0.000001 & 0.0000001 & 1 & 0.000001 & $1\times\rho$ & 0.0000001 & $0\times\rho$ & 1 \\
     20\%, NM (E) & 0.0000001 & 0.0000001 & 5 & 0.0000001 & $0\times\rho$ & 0.0000001 & $5\times\rho$ & 1 \\
     40\%, NM (E) & 0.0000001 & 0.0000001 & 1 & 0.000001 & $10\times\rho$ & 0.0000001 & $5\times\rho$ & 1 \\
    \bottomrule
  \end{tabular}
\end{table}

\subsection{TRMF and BTMF}

% {\color{red} make sure the values are correct, and how you tuned them.

Time lags of imputation and prediction for both TRMF and BTMF are set as $\mathcal{H}=\{1,2\}$ and $\mathcal{H}=\{1,...,6,I-2,...,I+3\}$, respectively. For prediction tasks, the low rank of PeMS data prediction is 20 while of Guangzhou/Electricity data prediction is 10. For imputation tasks, the low ranks are:
\begin{itemize}
    \item (\textbf{P}) data: 50 (for RM scenarios) and 10 (for NM scenarios) for both TRMF and BTMF.
    \item (\textbf{G}) data: 80 (for RM scenarios) and 10 (for NM scenarios) for both TRMF and BTMF.
    \item (\textbf{E}) data: 30 (for RM scenarios) for both TRMF and BTMF. For NM scenarios, we set 10, 30 for TRMF and BTMF, respectively.
\end{itemize}

% }

% \subsection{SLRTC}

% {\color{red} the $L$ does not make sense

% For tensor-based models, we set window length $L$ to 96 for PeMS (P) data, 144 for Guangzhou (G) data, and 24 for Electricity (E), respectively. We use the data of previous four days plus $\tau$ to build block Hankel tensors for all three data sets. The parameter $\alpha$ in the SLRTC model is given as: $\alpha=-0.005$ (P), $\alpha=-0.01$ (G), and $\alpha=-0.05$ (E).
% }

\section{Imputation/prediction performance}\label{imputation-prediction-example}

\begin{figure*}[!ht]
\centering
\subfigure[Sensor \#2.]{
    \centering
    \includegraphics[scale=0.44]{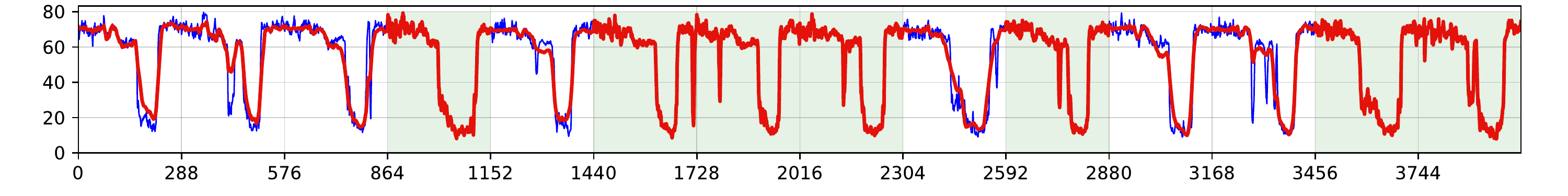}
}
\subfigure[Sensor \#3.]{
    \centering
    \includegraphics[scale=0.44]{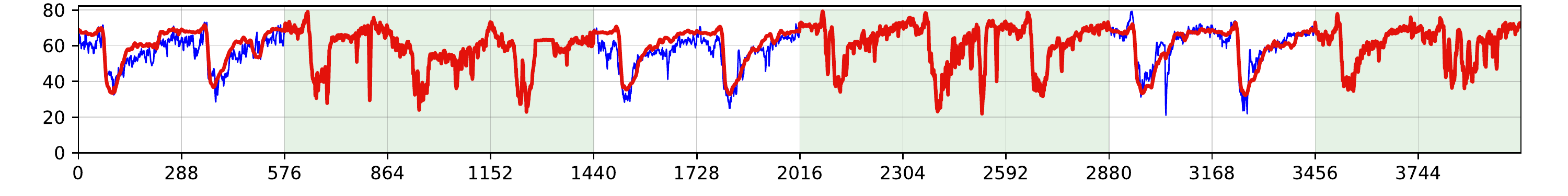}
}
\subfigure[Sensor \#4.]{
    \centering
    \includegraphics[scale=0.44]{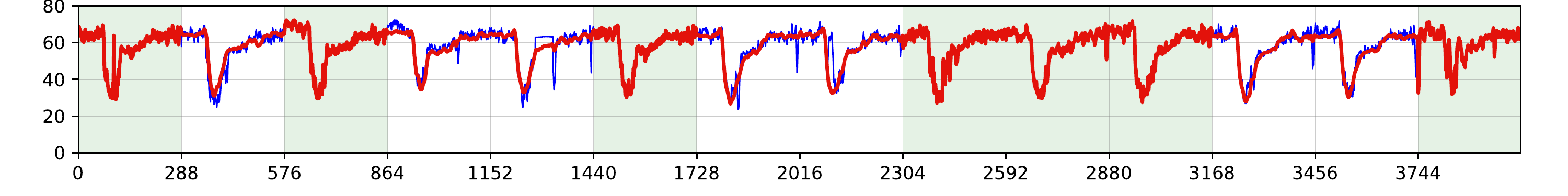}
}
\subfigure[Sensor \#5.]{
    \centering
    \includegraphics[scale=0.44]{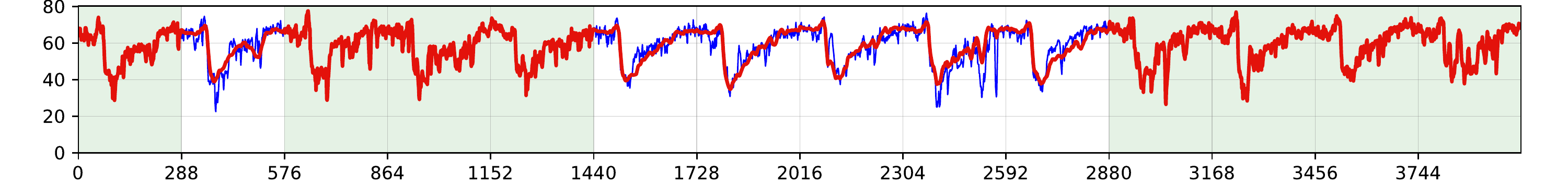}
}
\caption{Imputation of LATC-TNN on the PeMS data (first two weeks). We show the results from four sensors under 40\% NM pattern. The green panels indicate partially observed input data, the blue curves show the ground truth values, and the red curves show the recovered data.}
\label{pems_imputation}
\end{figure*}

\begin{figure*}[!ht]
\centering
\subfigure[Location \#1.]{
    \centering
    \includegraphics[scale=0.44]{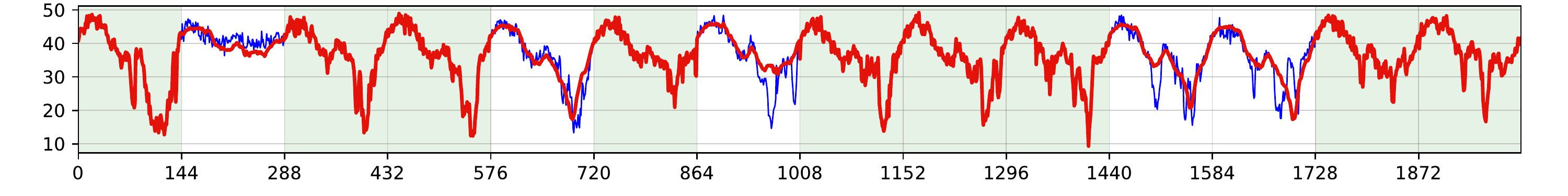}
}
\subfigure[Location \#2.]{
    \centering
    \includegraphics[scale=0.44]{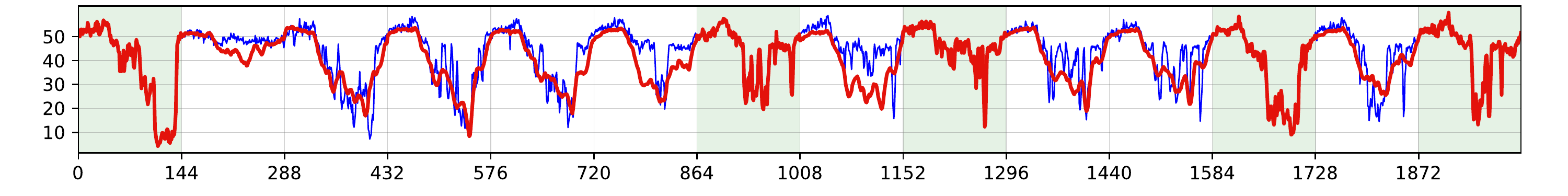}
}
\subfigure[Location \#3.]{
    \centering
    \includegraphics[scale=0.44]{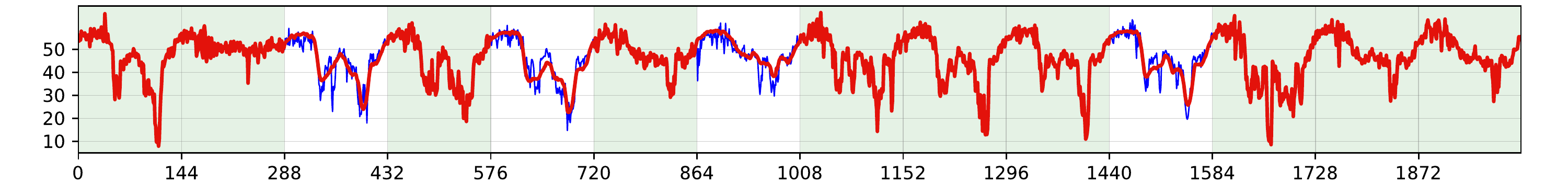}
}
\subfigure[Location \#4.]{
    \centering
    \includegraphics[scale=0.44]{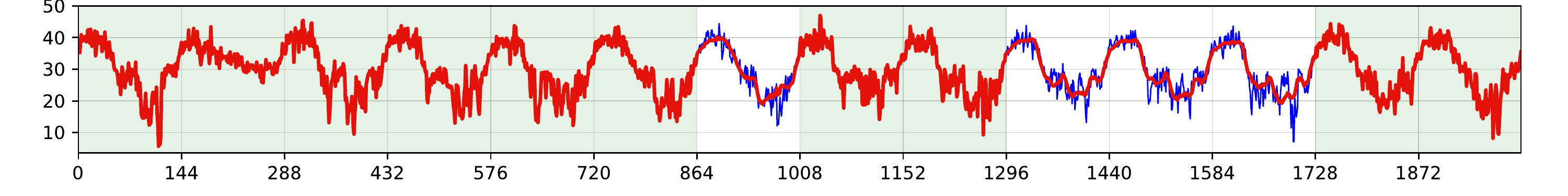}
}
\caption{Imputation of LATC-TNN on the Guangzhou data (first two weeks). We show the results from four locations under 40\% NM pattern. The green panels indicate partially observed input data, the blue curves show the ground truth values, and the red curves show the recovered data.}
\label{guangzhou_imputation}
\end{figure*}

\begin{figure*}[!ht]
\centering
\subfigure[Client \#2.]{
    \centering
    \includegraphics[scale=0.44]{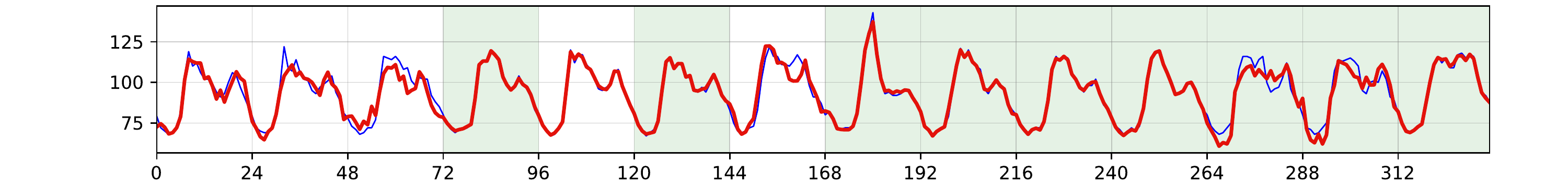}
}
\subfigure[Client \#4.]{
    \centering
    \includegraphics[scale=0.44]{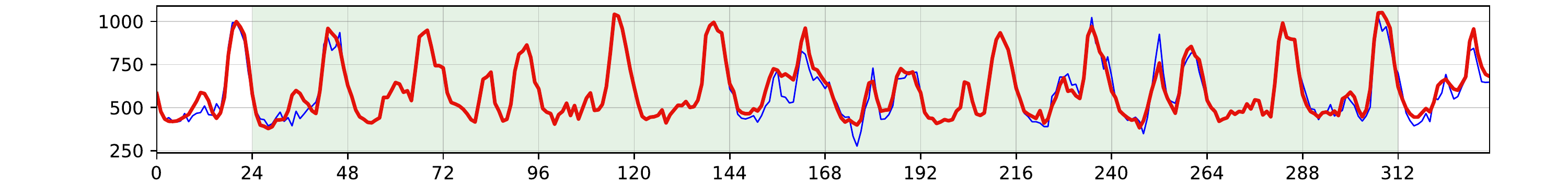}
}
\subfigure[Client \#5.]{
    \centering
    \includegraphics[scale=0.44]{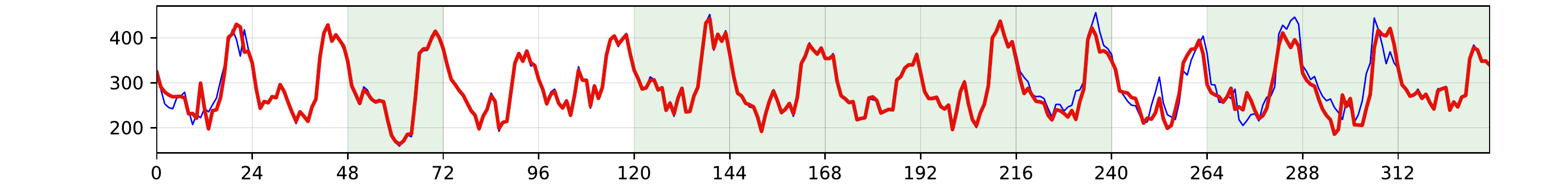}
}
\caption{Imputation of LATC-TNN on the Electricity data (first two weeks). We show the results from three clients under 40\% NM pattern. The green panels indicate partially observed input data, the blue curves show the ground truth values, and the red curves show the recovered data.}
\label{electricity_imputation}
\end{figure*}

\begin{figure*}[!ht]
\centering
\subfigure[Sensor \#2 (40\%, RM).]{
    \centering
    \includegraphics[scale=0.44]{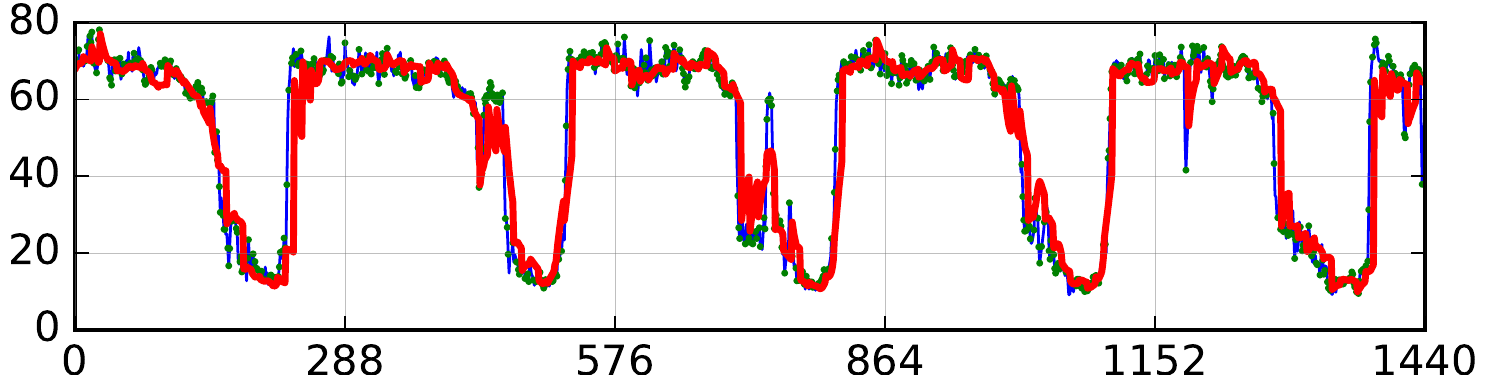}
}
\subfigure[Sensor \#2 (40\%, NM).]{
    \centering
    \includegraphics[scale=0.44]{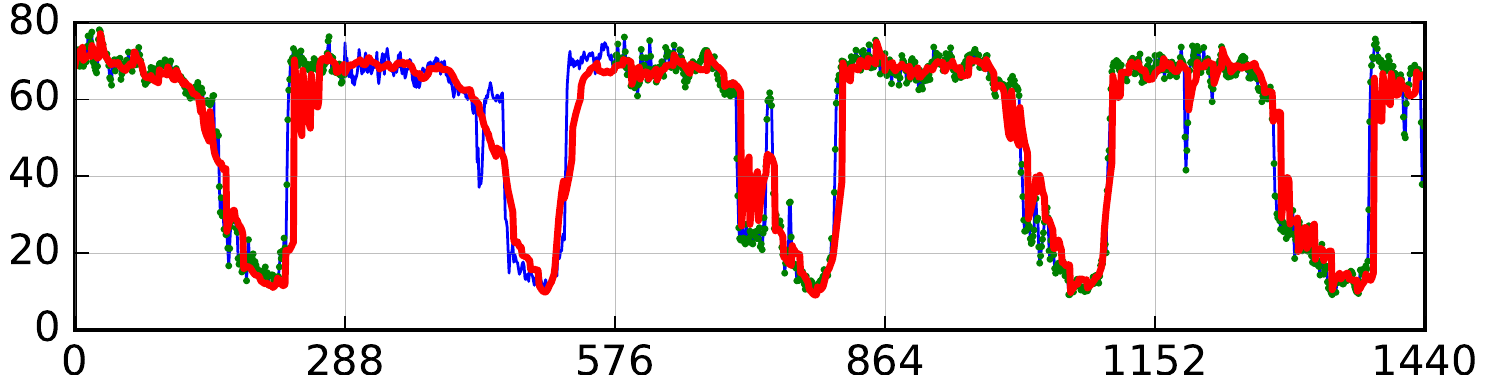}
}
\subfigure[Sensor \#3 (40\%, RM).]{
    \centering
    \includegraphics[scale=0.44]{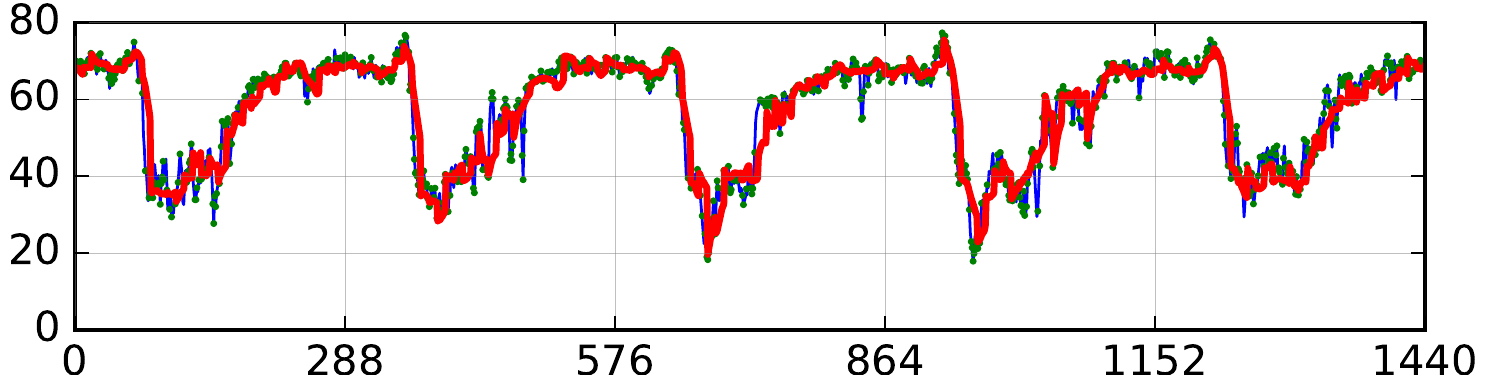}
}
\subfigure[Sensor \#3 (40\%, NM).]{
    \centering
    \includegraphics[scale=0.44]{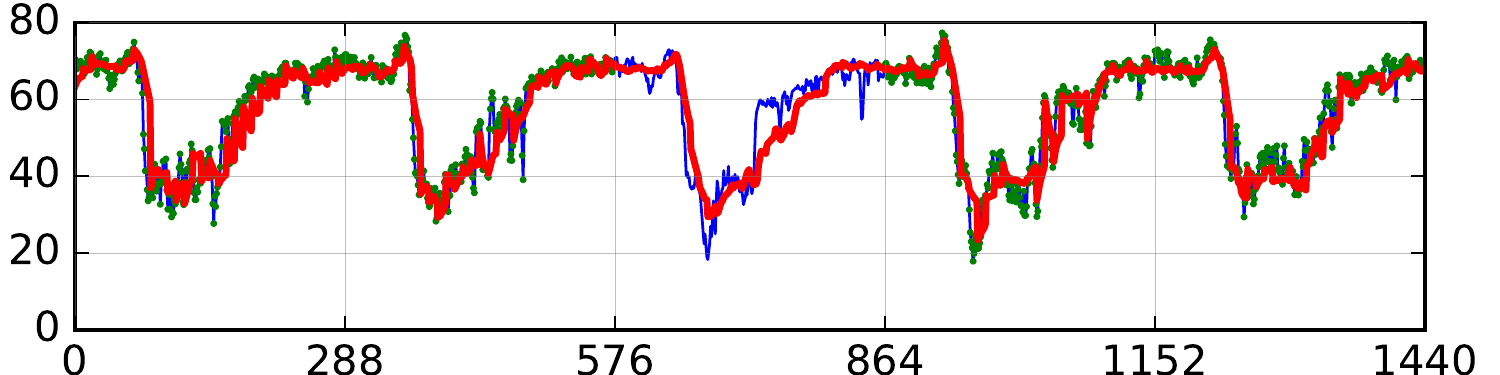}
}
\subfigure[Sensor \#4 (40\%, RM).]{
    \centering
    \includegraphics[scale=0.44]{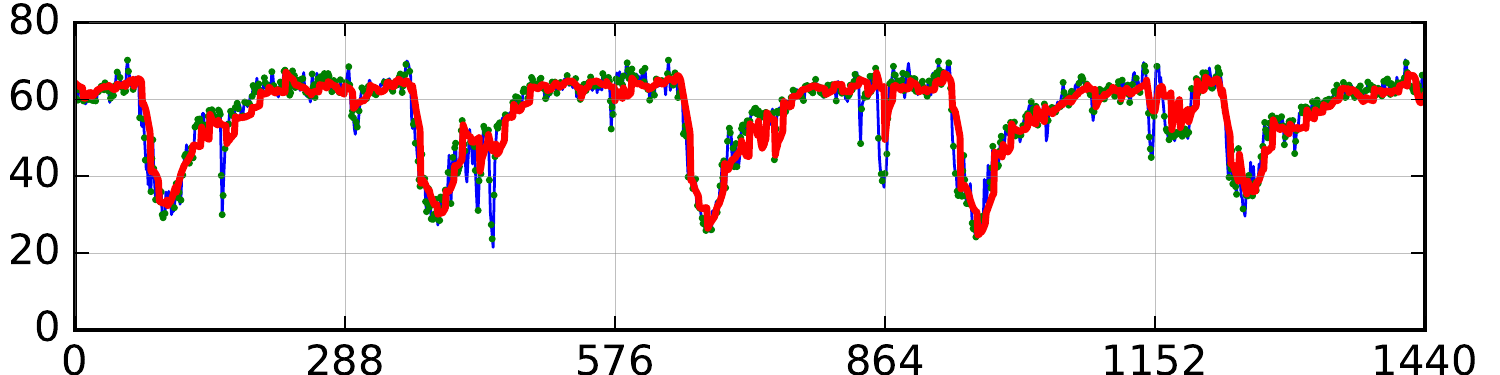}
}
\subfigure[Sensor \#4 (40\%, NM).]{
    \centering
    \includegraphics[scale=0.44]{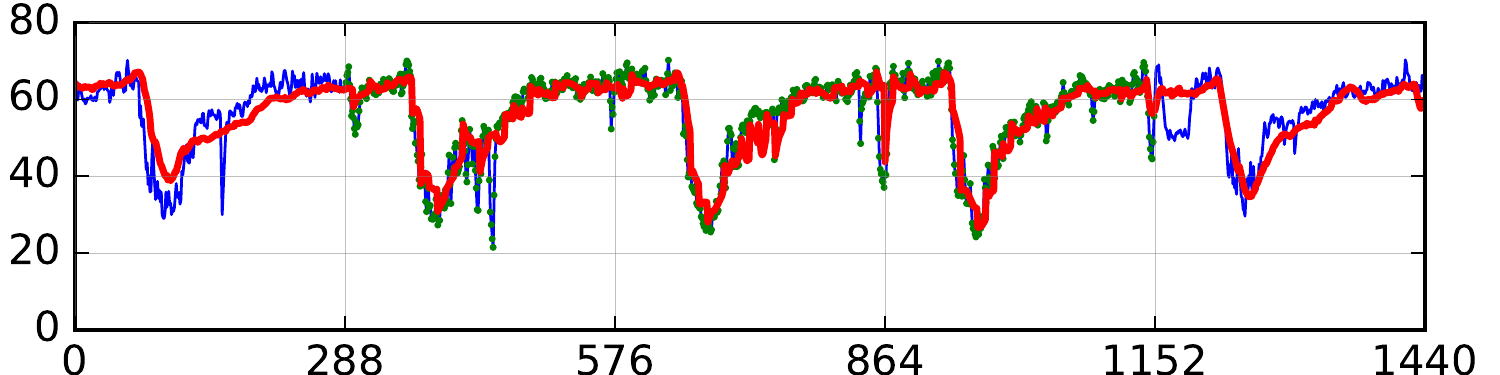}
}
\subfigure[Sensor \#5 (40\%, RM).]{
    \centering
    \includegraphics[scale=0.44]{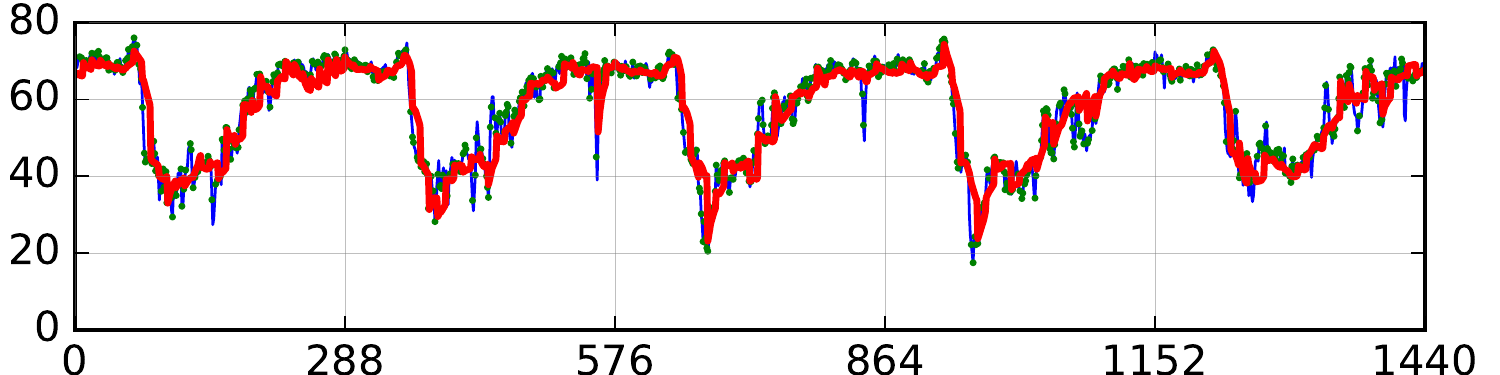}
}
\subfigure[Sensor \#5 (40\%, NM).]{
    \centering
    \includegraphics[scale=0.44]{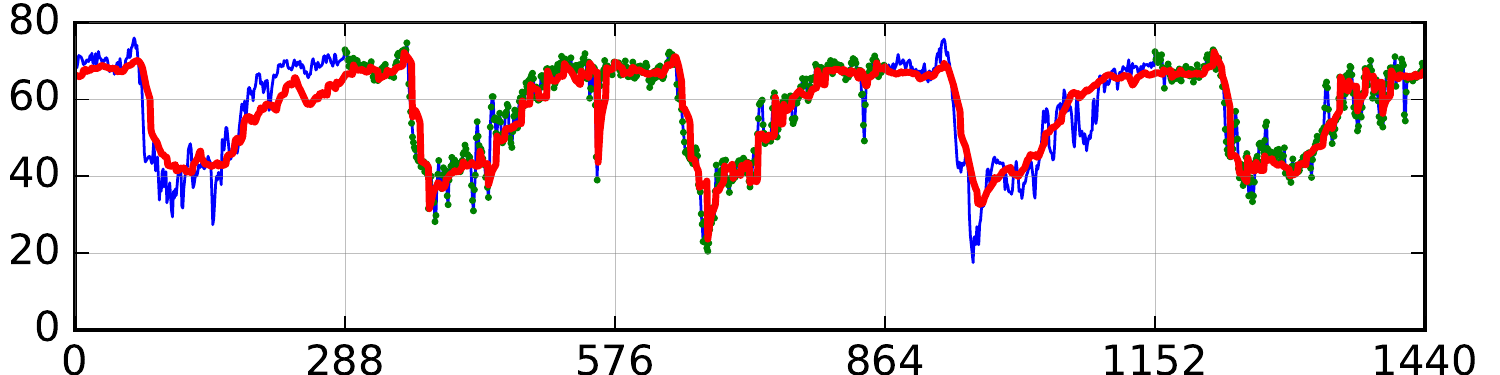}
}
\caption{Rolling prediction of LATC-TNN on the PeMS data ($\tau=9$ and $S=160$, 1440 time points in total). We show the results from four sensors under different missing scenarios. The green dots represent partially observed input data, the blue curves show the ground truth values, and the red curves show the prediction.}
\label{pems_forecasting}
\end{figure*}

\begin{figure*}[!ht]
\centering
\subfigure[Location \#1 (40\%, RM).]{
    \centering
    \includegraphics[scale=0.44]{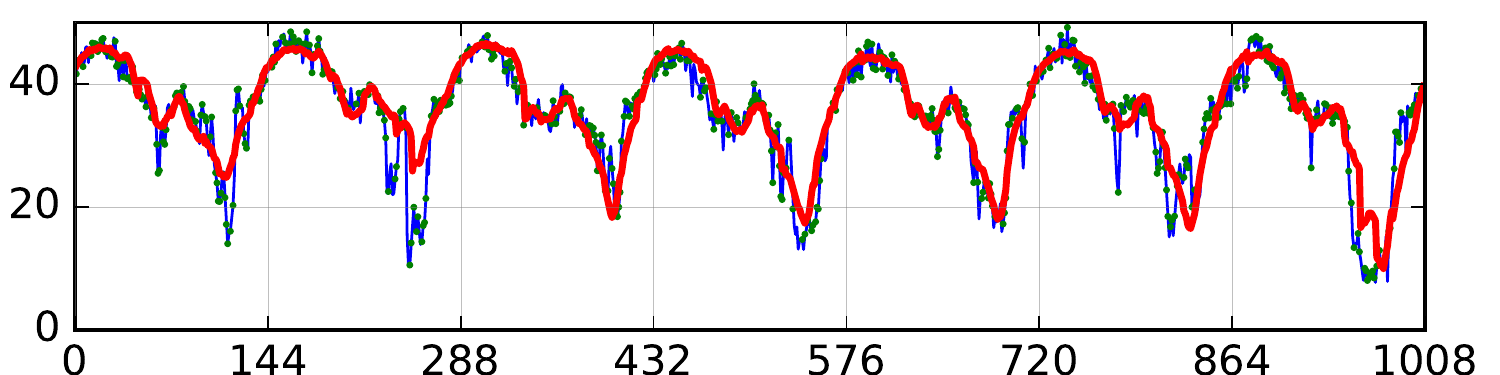}
}
\subfigure[Location \#1 (40\%, NM).]{
    \centering
    \includegraphics[scale=0.44]{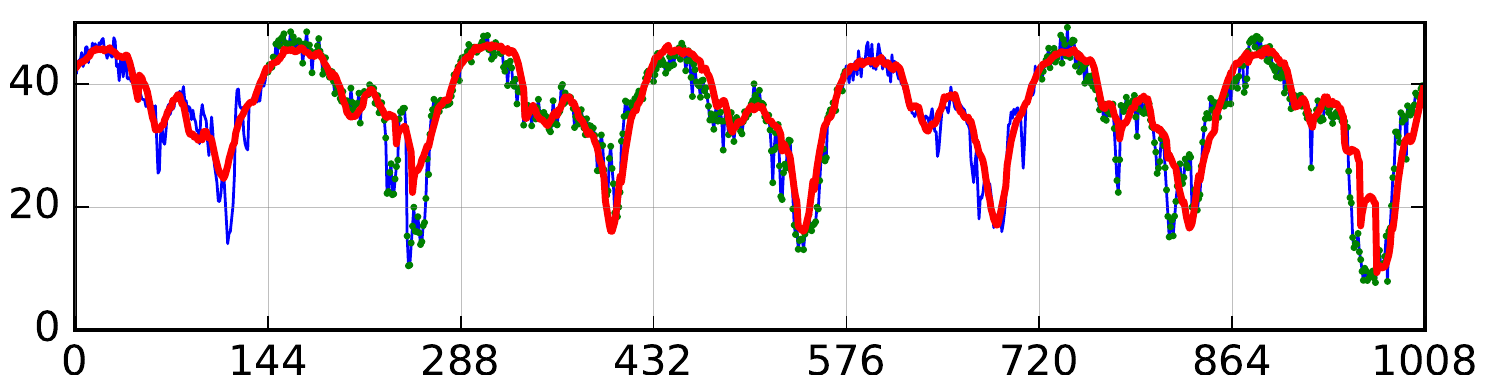}
}
\subfigure[Location \#2 (40\%, RM).]{
    \centering
    \includegraphics[scale=0.44]{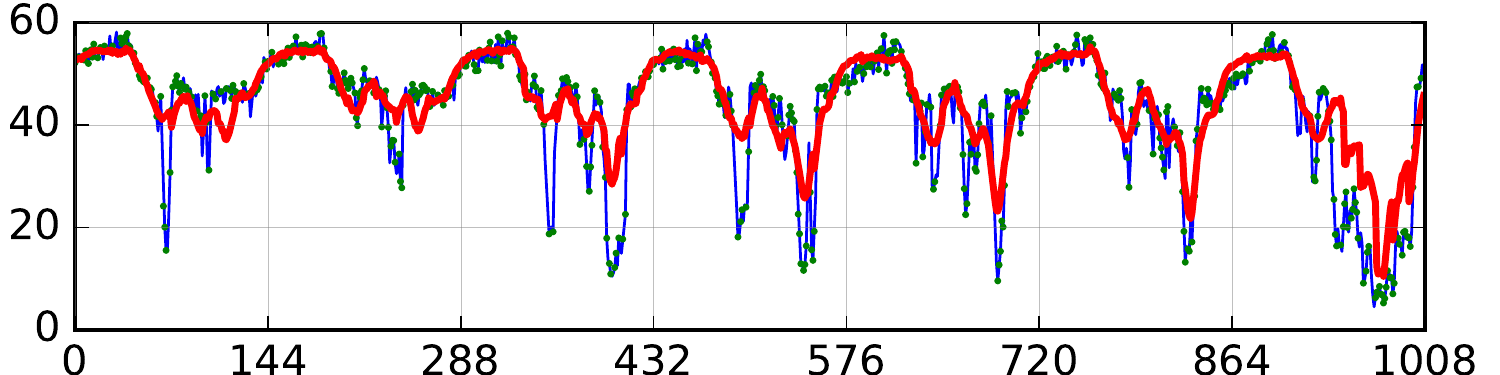}
}
\subfigure[Location \#2 (40\%, NM).]{
    \centering
    \includegraphics[scale=0.44]{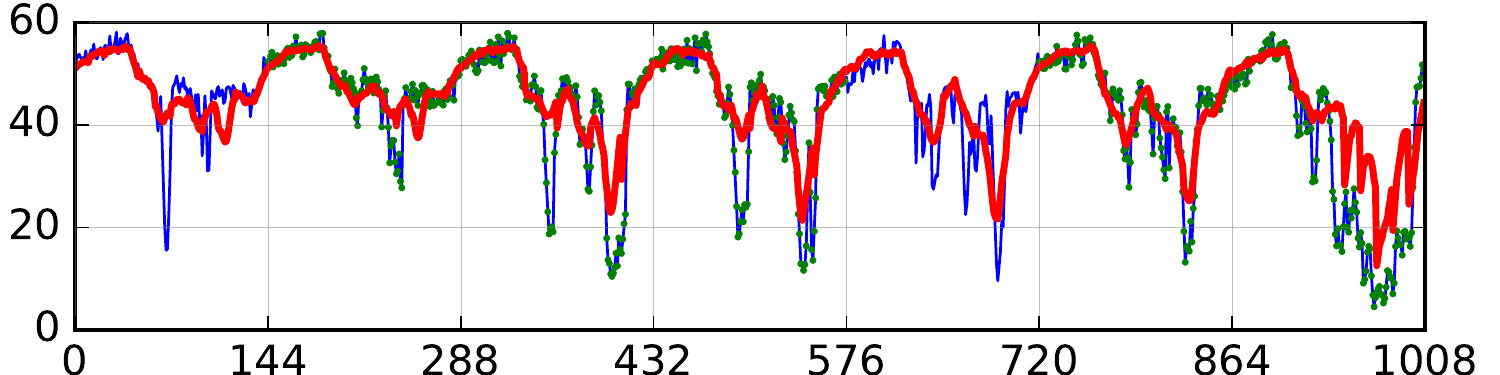}
}
\subfigure[Location \#3 (40\%, RM).]{
    \centering
    \includegraphics[scale=0.44]{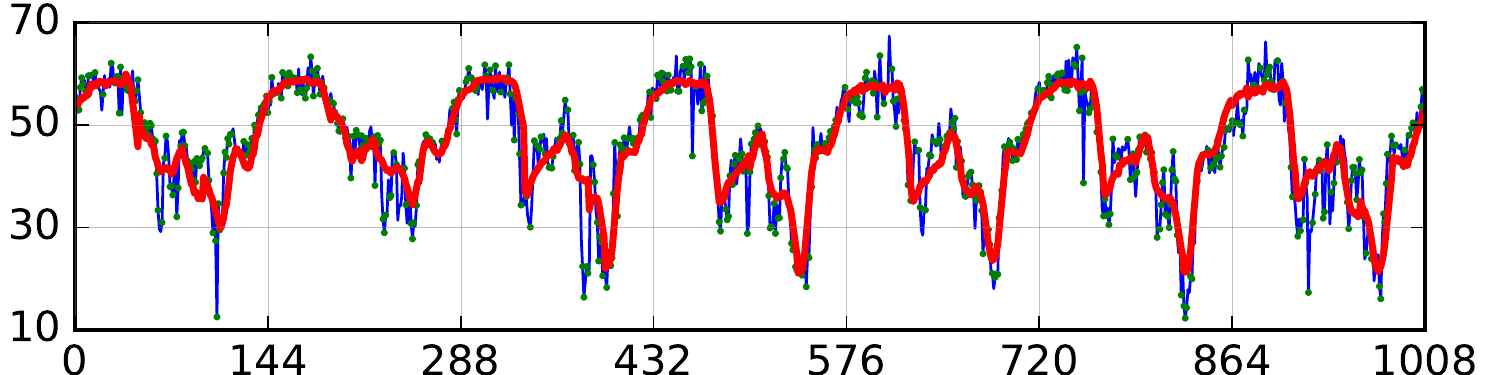}
}
\subfigure[Location \#3 (40\%, NM).]{
    \centering
    \includegraphics[scale=0.44]{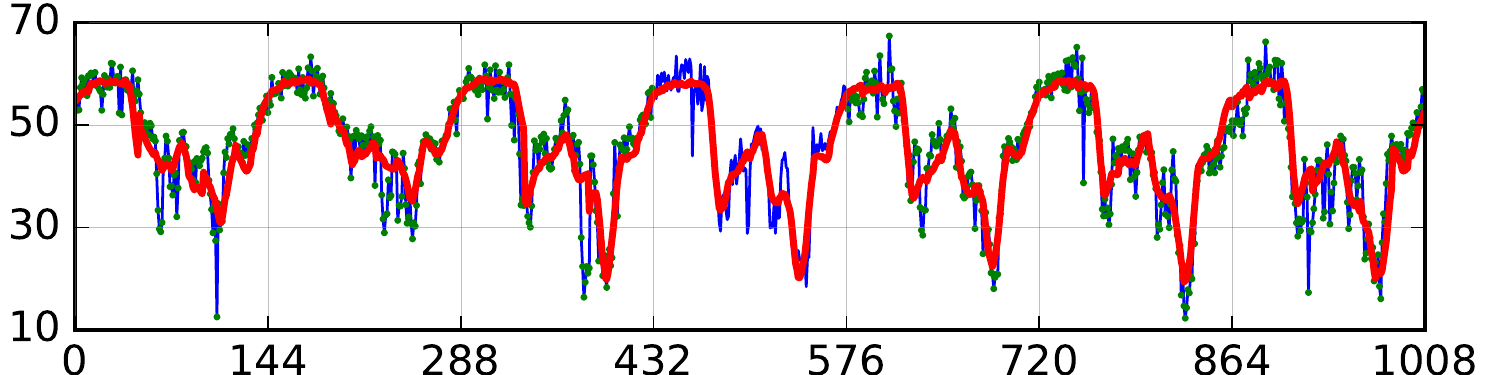}
}
\subfigure[Location \#4 (40\%, RM).]{
    \centering
    \includegraphics[scale=0.44]{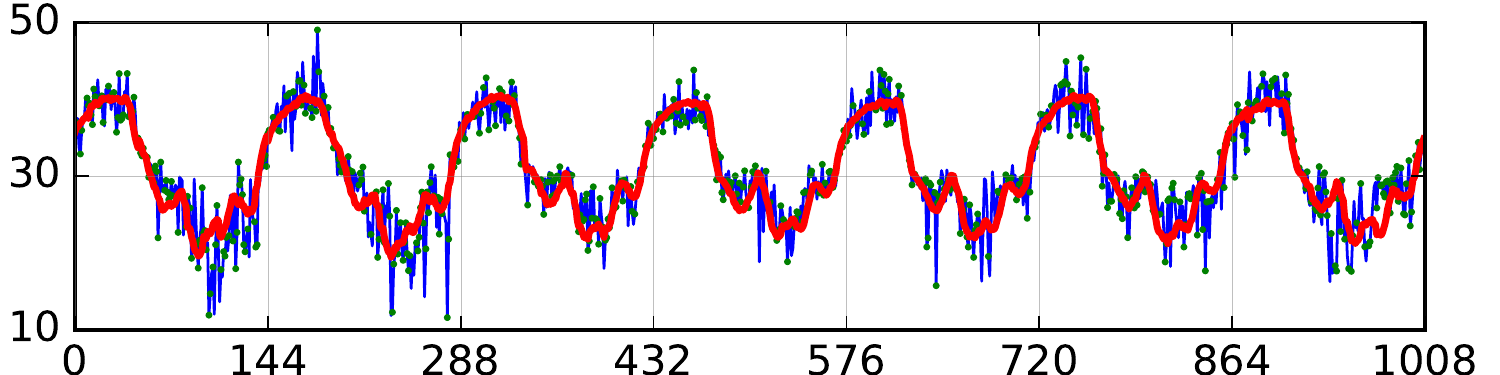}
}
\subfigure[Location \#4 (40\%, NM).]{
    \centering
    \includegraphics[scale=0.44]{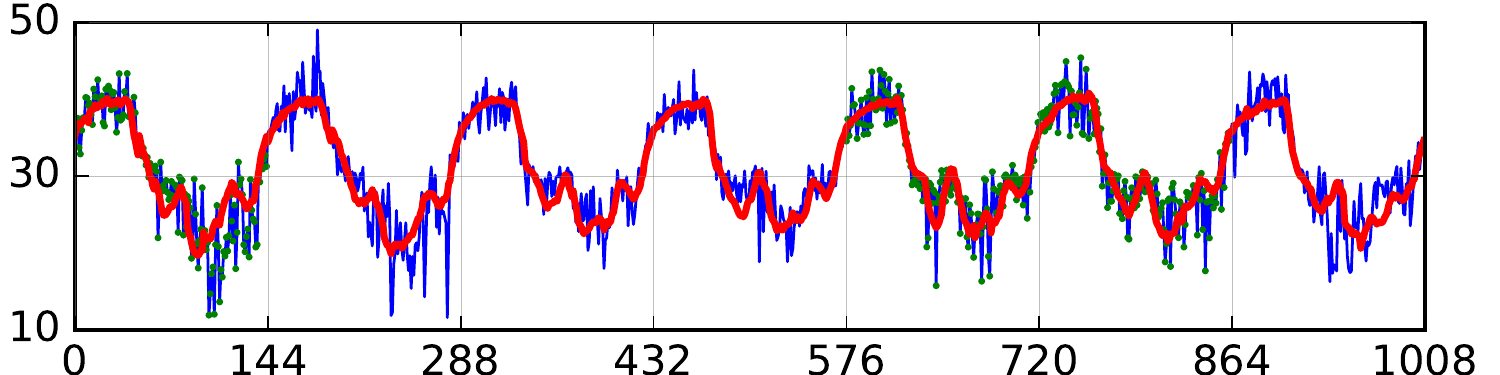}
}
\caption{Rolling prediction of LATC-TNN on the Guangzhou data ($\tau=12$ and $S=84$, 1008 time points in total). We show the results from four locations under different missing scenarios. The green dots represent partially observed input data, the blue curves show the ground truth values, and the red curves show the prediction.}
\label{guangzhou_forecasting}
\end{figure*}

\begin{figure*}[!ht]
\centering
\subfigure[Client \#2 (40\%, RM).]{
    \centering
    \includegraphics[scale=0.44]{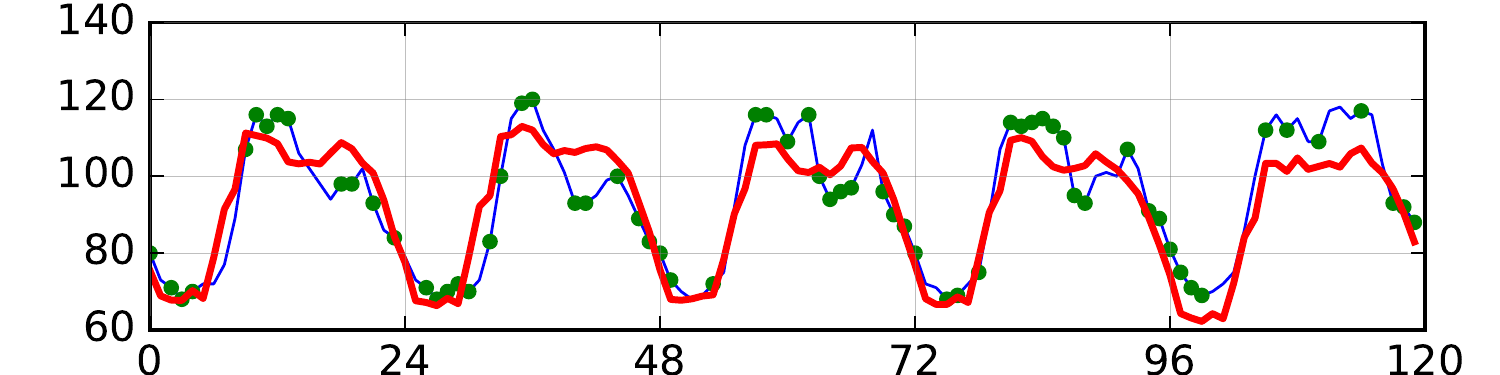}
}
\subfigure[Client \#2 (40\%, NM).]{
    \centering
    \includegraphics[scale=0.44]{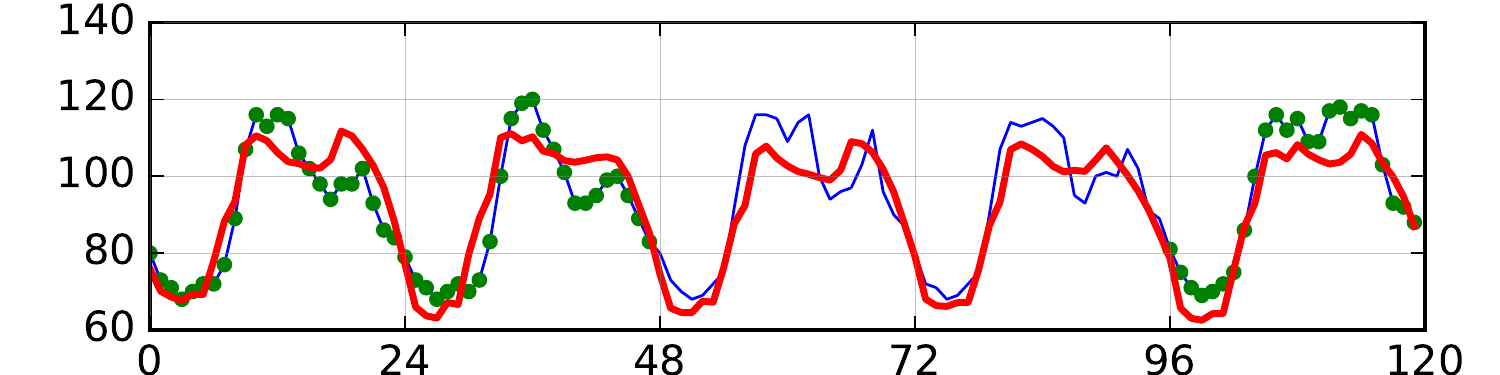}
}
\subfigure[Client \#4 (40\%, RM).]{
    \centering
    \includegraphics[scale=0.44]{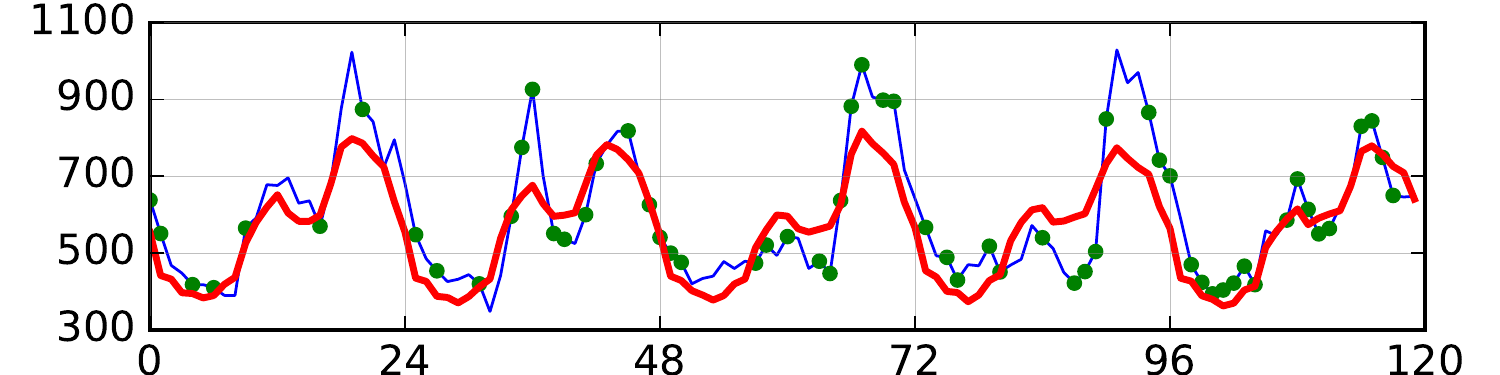}
}
\subfigure[Client \#4 (40\%, NM).]{
    \centering
    \includegraphics[scale=0.44]{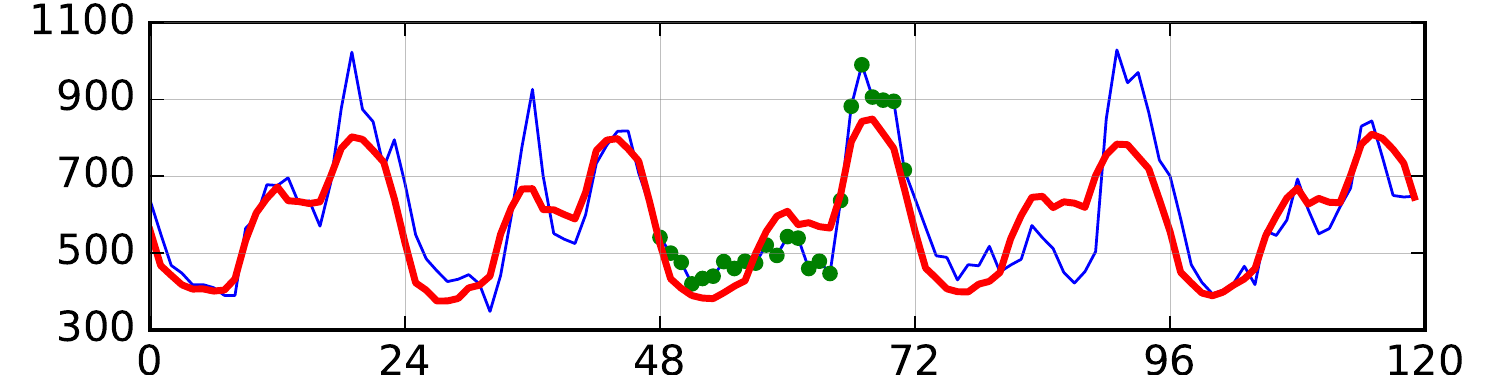}
}
\subfigure[Client \#5 (40\%, RM).]{
    \centering
    \includegraphics[scale=0.44]{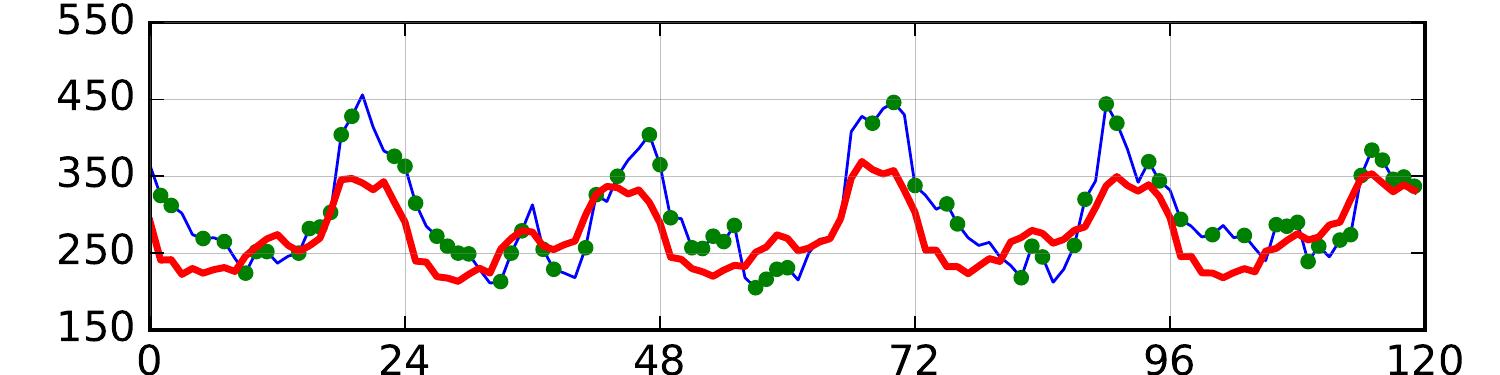}
}
\subfigure[Client \#5 (40\%, NM).]{
    \centering
    \includegraphics[scale=0.44]{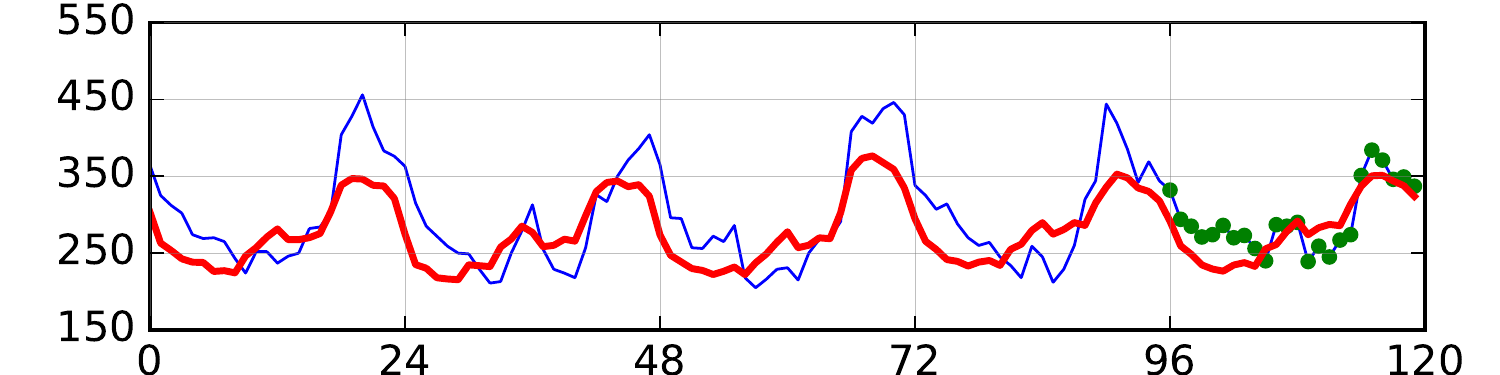}
}
\caption{Rolling prediction of LATC-TNN on the Electricity data ($\tau=6$ and $S=20$, 120 time points in total). We show the results from three clients under different missing scenarios. The green dots represent partially observed input data, the blue curves show the ground truth values, and the red curves show the prediction.}
\label{Electricity_forecasting}
\end{figure*}

In this section, we provide some visualizations to demonstrate the performance of LATC. Figures~\ref{pems_imputation}, \ref{guangzhou_imputation}, and \ref{electricity_imputation} show the imputation performance of LATC-TNN on different data sets under the 40\% non-random missing (NM) scenarios. In all the three figures, the green panels represent the observed data, and the white panels correspond to missing values to impute. The blue curves are the ground-truth, and the red curves show the recovered matrix $\hat{\boldsymbol{X}}$. Figures~\ref{pems_forecasting}, \ref{guangzhou_forecasting}, and \ref{Electricity_forecasting} show the prediction performance of LATC-TNN on different data sets under two missing scenarios (i.e., 40\% random missing and 40\% non-random missing). These figures only show the final predicted time windows. The prediction is performed by a rolling-window approach: in each step, we predict a length-$\tau$ time window (see Algorithm~\ref{predictor}).

\section{Derivation detail}

In this section, we provide detailed derivation of some optimization problems in LATC.

\subsection{Updating the variable $\boldsymbol{\mathcal{X}}_{k},k=1,2,3$}

For the optimization problem in Eq.~\eqref{tensor_sub1}, we can first write it as follows,
\begin{equation}
\begin{aligned}
    \boldsymbol{\mathcal{X}}_{k}^{l+1}:=&\operatorname{arg}\min_{\boldsymbol{\mathcal{X}}}~\alpha_k\|\boldsymbol{\mathcal{X}}_{(k)}\|_{*}+\frac{\rho}{2}\|\mathcal{Q}^{-1}(\boldsymbol{\mathcal{X}})-\boldsymbol{Z}^{l}\|_{F}^{2}+\big\langle\mathcal{Q}^{-1}(\boldsymbol{\mathcal{X}})-\boldsymbol{Z}^{l},\mathcal{Q}^{-1}(\boldsymbol{\mathcal{T}}_{k}^{l})\big\rangle, \\
    =&\operatorname{arg}\min_{\boldsymbol{\mathcal{X}}}~\alpha_k\|\boldsymbol{\mathcal{X}}_{(k)}\|_{*}+\frac{\rho}{2}\big\langle\mathcal{Q}^{-1}(\boldsymbol{\mathcal{X}})-\boldsymbol{Z}^{l},\mathcal{Q}^{-1}(\boldsymbol{\mathcal{X}})-\boldsymbol{Z}^{l}\big\rangle+\big\langle\mathcal{Q}^{-1}(\boldsymbol{\mathcal{X}}),\mathcal{Q}^{-1}(\boldsymbol{\mathcal{T}}_{k}^{l})\big\rangle \\
    =&\operatorname{arg}\min_{\boldsymbol{\mathcal{X}}}~\alpha_k\|\boldsymbol{\mathcal{X}}_{(k)}\|_{*}+\frac{\rho}{2}\big\langle\mathcal{Q}^{-1}(\boldsymbol{\mathcal{X}}),\mathcal{Q}^{-1}(\boldsymbol{\mathcal{X}})\big\rangle-\big\langle\mathcal{Q}^{-1}(\boldsymbol{\mathcal{X}}),\rho\boldsymbol{Z}^{l}-\mathcal{Q}^{-1}(\boldsymbol{\mathcal{T}}_{k}^{l})\big\rangle \\
    =&\operatorname{arg}\min_{\boldsymbol{\mathcal{X}}}~\alpha_k\|\boldsymbol{\mathcal{X}}_{(k)}\|_{*}+\frac{\rho}{2}\big\langle\mathcal{Q}^{-1}(\boldsymbol{\mathcal{X}})-\boldsymbol{Z}^{l}+\frac{1}{\rho}\mathcal{Q}^{-1}(\boldsymbol{\mathcal{T}}_{k}^{l}),\mathcal{Q}^{-1}(\boldsymbol{\mathcal{X}})-\boldsymbol{Z}^{l}+\frac{1}{\rho}\mathcal{Q}^{-1}(\boldsymbol{\mathcal{T}}_{k}^{l})\big\rangle \\
    =&\operatorname{arg}\min_{\boldsymbol{\mathcal{X}}}~\alpha_k\|\boldsymbol{\mathcal{X}}_{(k)}\|_{*}+\frac{\rho}{2}\|\mathcal{Q}^{-1}(\boldsymbol{\mathcal{X}})-\boldsymbol{Z}^{l}+\mathcal{Q}^{-1}(\boldsymbol{\mathcal{T}}_{k}^{l})/\rho\|_{F}^{2} \\
    =&\operatorname{arg}\min_{\boldsymbol{\mathcal{X}}}~\alpha_k\|\boldsymbol{\mathcal{X}}_{(k)}\|_{*}+\frac{\rho}{2}\|\boldsymbol{\mathcal{X}}_{(k)}-\mathcal{Q}(\boldsymbol{Z}^{l})_{(k)}+\boldsymbol{\mathcal{T}}_{k(k)}^{l}/\rho\|_{F}^{2}, \\
\end{aligned}
\end{equation}
then applying Lemma~\ref{lemma2}, we can therefore obtain the optimal solution to the variable $\boldsymbol{\mathcal{X}}$ as
\begin{equation}
    \boldsymbol{\mathcal{X}}_{k}^{l+1}:=\operatorname{fold}_{k}\left(\mathcal{D}_{\alpha_k/\rho}(\mathcal{Q}(\boldsymbol{Z}^{l})_{(k)}-\boldsymbol{\mathcal{T}}_{k(k)}^{l}/\rho)\right).
\end{equation}

Similarly, we can write the closed-form solution to Eq.~\eqref{tnn_sub1} with TNN minimization.

\subsection{Updating the variable $\boldsymbol{Z}$}

For the optimization problem in Eq.~\eqref{tensor_sub2}, we have
\begin{equation}
\begin{aligned}
    \boldsymbol{Z}^{l+1}_{[:h_d]}:=&
    \operatorname{arg}\min_{\boldsymbol{Z}}~\sum_{k}\left(\frac{\rho}{2}\|\mathcal{Q}^{-1}(\boldsymbol{\mathcal{X}}_{k}^{l+1})_{[:h_d]}-\boldsymbol{Z}\|_{F}^{2}
    +\big\langle\mathcal{Q}^{-1}(\boldsymbol{\mathcal{X}}_{k}^{l+1})_{[:h_d]}-\boldsymbol{Z},\mathcal{Q}^{-1}(\boldsymbol{\mathcal{T}}_{k}^{l})_{[:h_d]}\big\rangle\right) \\
    =&\operatorname{arg}\min_{\boldsymbol{Z}}~\sum_{k}\left(\frac{\rho}{2}\big\langle\boldsymbol{Z},\boldsymbol{Z}\big\rangle-\rho\big\langle\boldsymbol{Z},\mathcal{Q}^{-1}(\boldsymbol{\mathcal{X}}_{k}^{l+1}+\boldsymbol{\mathcal{T}}_{k}^{l}/\rho)_{[:h_d]}\big\rangle\right) \\
    =&\operatorname{arg}\min_{\boldsymbol{Z}}~\frac{\rho}{2}\sum_{k}\big\langle\boldsymbol{Z}-\mathcal{Q}^{-1}(\boldsymbol{\mathcal{X}}_{k}^{l+1}+\boldsymbol{\mathcal{T}}_{k}^{l}/\rho)_{[:h_d]},\boldsymbol{Z}-\mathcal{Q}^{-1}(\boldsymbol{\mathcal{X}}_{k}^{l+1}+\boldsymbol{\mathcal{T}}_{k}^{l}/\rho)_{[:h_d]}\big\rangle \\
    =&\frac{1}{3}\sum_{k}\mathcal{Q}^{-1}(\boldsymbol{\mathcal{X}}_{k}^{l+1}+\boldsymbol{\mathcal{T}}_{k}^{l}/\rho)_{[:h_d]}. \\
\end{aligned}
\end{equation}
% and this problem has a closed-form least square solution as

For the optimization problem in Eq.~\eqref{tensor_sub3}, we have
\begin{equation}
\begin{aligned}
    \boldsymbol{Z}^{l+1}_{[h_d+1:]}&
    \begin{aligned}[t]
    &:=\operatorname{arg}\min_{\boldsymbol{Z}}~\sum_{m}\frac{\lambda}{2}\|\boldsymbol{z}_{m,[h_d+1:]}-\boldsymbol{Q}_{m}\boldsymbol{a}_{m}\|_{2}^{2}+\\ &\sum_{k}\left(\frac{\rho}{2}\|\mathcal{Q}^{-1}(\boldsymbol{\mathcal{X}}_{k}^{l+1})_{[h_d+1:]}-\boldsymbol{Z}\|_{F}^{2}+\big\langle\mathcal{Q}^{-1}(\boldsymbol{\mathcal{X}}_{k}^{l+1})_{[h_d+1:]}-\boldsymbol{Z},\mathcal{Q}^{-1}(\boldsymbol{\mathcal{T}}_{k}^{l})_{[h_d+1:]}\big\rangle\right)\\
    \end{aligned}
\end{aligned}
\end{equation}
\begin{equation}
\begin{aligned}
    \Rightarrow\boldsymbol{z}_{m,[h_d+1:]}:=&\operatorname{arg}\min_{\boldsymbol{z}}~\frac{\lambda}{2}\|\boldsymbol{z}-\boldsymbol{Q}_{m}\boldsymbol{a}_{m}\|_{F}^{2}+\frac{\rho}{2}\|\boldsymbol{z}-\frac{1}{3}\sum_{k}\mathcal{Q}^{-1}(\boldsymbol{\mathcal{X}}_{k}^{l+1}+\boldsymbol{\mathcal{T}}_{k}^{l}/\rho)_{m,[h_d+1:]}\|_{F}^{2} \\
    =&\frac{\lambda}{\rho+\lambda}\boldsymbol{Q}_{m}\boldsymbol{a}_{m}^{l}+\frac{1}{3(\rho+\lambda)}\sum_{k}\mathcal{Q}^{-1}(\rho\boldsymbol{\mathcal{X}}_{k}^{l+1}+\boldsymbol{\mathcal{T}}_{k}^{l})_{m,[h_d+1:]}. \\
\end{aligned}
\end{equation}

% \section{Comparison with Other SOTA Models}

% Prediction MSE ($\times 10^{-4}$) for the hourly \texttt{traffic} data (forecast the last 7 days with $\tau=24$):

% \begin{itemize}
%     \item VAR: 51
%     \item GARCH: 33
%     \item Vec-LSTM-ind: 6.5
%     \item Vec-LSTM-ind-scaling: 6.3
%     \item Vec-LSTM-lowrank-Copula: 15
%     \item GP-scaling: 5.2
%     \item GP-Copula: 6.9
%     \item \textbf{Our LATC-NN: 4.4}
%     \item \textbf{Our LATC-TNN: 4.0}
% \end{itemize}
% Note that these results are from Table 6 in \url{https://arxiv.org/pdf/1910.03002.pdf}.

% Precition RSE for hourly \texttt{traffic} data (forecast the last ? days with $\tau=24$):

% \begin{itemize}
%     \item AR: 0.6293
%     \item LRidge: 0.6025
%     \item LSVR: 0.5909
%     \item TRMF: 0.6442
%     \item GP: 0.5995
%     \item VARMLP: 0.6146
%     \item RNN-GRU: 0.5633
%     \item LST-Skip: 0.4973
%     \item LST-Attn: 0.5300
%     \item \textbf{Our LATC-NN: 0.5596}
%     \item \textbf{Our LATC-TNN: 0.5201}
% \end{itemize}
% Note that these results are from Table 2 in \url{https://dl.acm.org/doi/pdf/10.1145/3209978.3210006}.

\end{document}